\documentclass[10pt,twocolumn,letterpaper]{article}

\usepackage{iccv}
\usepackage{times}
\usepackage{epsfig}
\usepackage{graphicx}
\usepackage{amsmath}
\usepackage{amssymb}

\usepackage{float}
\usepackage{graphicx}
\usepackage{amsmath}
\usepackage{amssymb}
\usepackage{booktabs}
\usepackage{multirow, makecell} 

\usepackage[table,xcdraw]{xcolor} 
\newcommand{\gr}{\rowcolor[gray]{.95}} 

\usepackage[para]{footmisc} 

\usepackage{pifont}
\newcommand{\cmark}{\ding{51}}
\newcommand{\xmark}{\ding{55}}

\usepackage[pagebackref=true,breaklinks=true,letterpaper=true,colorlinks,bookmarks=false]{hyperref}

\iccvfinalcopy 

\def\iccvPaperID{2479} 

\ificcvfinal\fi

\begin{document}

\title{CRN: Camera Radar Net for Accurate, Robust, Efficient 3D Perception}

\author{
Youngseok Kim$^1$ \hspace{5pt} Juyeb Shin$^1$ \hspace{5pt} Sanmin Kim$^1$ \hspace{5pt} In-Jae Lee$^1$ \hspace{5pt} Jun Won Choi$^2$ \hspace{5pt} Dongsuk Kum$^1$ \\
$^1$KAIST \hspace{10pt} $^2$Hanyang University\\
{\tt\small \{youngseok.kim, juyebshin, sanmin.kim, injaelee, dskum\}@kaist.ac.kr, junwchoi@hanyang.ac.kr} 
}

\maketitle
\ificcvfinal\thispagestyle{empty}\fi

\begin{abstract}
Autonomous driving requires an accurate and fast 3D perception system that includes 3D object detection, tracking, and segmentation.
Although recent low-cost camera-based approaches have shown promising results, they are susceptible to poor illumination or bad weather conditions and have a large localization error.
Hence, fusing camera with low-cost radar, which provides precise long-range measurement and operates reliably in all environments, is promising but has not yet been thoroughly investigated.
In this paper, we propose Camera Radar Net (CRN), a novel camera-radar fusion framework that generates a semantically rich and spatially accurate bird's-eye-view (BEV) feature map for various tasks.
To overcome the lack of spatial information in an image, we transform perspective view image features to BEV with the help of sparse but accurate radar points.
We further aggregate image and radar feature maps in BEV using multi-modal deformable attention designed to tackle the spatial misalignment between inputs.
CRN with real-time setting operates at 20 FPS while achieving comparable performance to LiDAR detectors on nuScenes, and even outperforms at a far distance on $100m$ setting.
Moreover, CRN with offline setting yields 62.4\% NDS, 57.5\% mAP on nuScenes test set and ranks first among all camera and camera-radar 3D object detectors.
\end{abstract}

\section{Introduction}\label{sec:intro}








Accurate and robust 3D perception system is crucial for many applications, such as autonomous driving and mobile robot.
For efficient 3D perception, obtaining a reliable bird's eye view (BEV) feature map from sensor inputs is necessary since various downstream tasks can be operated on BEV space (\textit{e.g.}, object detection \& tracking~\cite{yin2021center}, BEV segmentation~\cite{zhou2022cross}, HD map generation~\cite{shin2023instagram}, trajectory prediction~\cite{hu2021fiery}, and motion planning~\cite{philion2020lift}).
Another important ingredient for deploying 3D perception to the real world is to build a system that relies less on LiDAR disadvantaged from high-cost, high-maintenance, and low-reliability.
Apart from the drawbacks of LiDAR, 3D perception system is required to identify semantic information on the road (\textit{e.g.}, traffic lights, road sign) that can be easily leveraged by camera.
In addition to the need for rich semantic information, detecting distant objects is essential, and this can be benefited from radar.

\begin{figure}[t]
\begin{center}
\includegraphics[width=0.96\columnwidth]{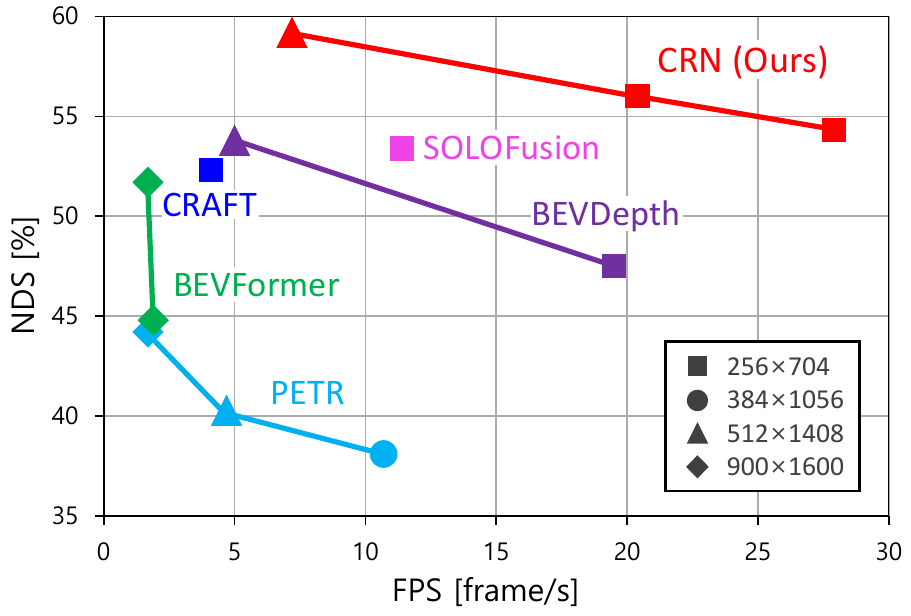}
\end{center}
\vspace{-10pt}
\caption{
    FPS vs. accuracy on nuScenes \texttt{val} set. 
    We show that fusing radar can significantly boost camera-only method with marginal computational cost.
    CRN outperforms all methods with much faster speed.
    See Table~\ref{table:3D det val set} and Fig.~\ref{fig:analysis_speed} for more details.
}
\label{fig:intro}
\vspace{-8pt}
\end{figure}

Recently, camera-based 3D perception in BEV~\cite{huang2021bevdet, philion2020lift, reading2021categorical} has drawn great attention.
Thanks to rich semantic information in dense image pixels, camera approaches can distinguish objects even at a far distance.
Despite the advantage of cameras, localizing the accurate position of objects from a monocular image is naturally a challenging ill-posed problem.
Moreover, cameras can be significantly affected by illumination conditions (\textit{e.g.}, glare, low-contrast, or low-lighting) due to the nature of the passive sensor.
To address this, \textit{we aim to generate a BEV feature map using a camera with the help of a cost-effective range sensor, radar}.

Radar has advantages not only in cost but also in high-reliability, long-range perception (up to 200$m$ for typical automotive radar~\cite{ars408}), robustness in various conditions (\textit{e.g.}, snow, fog, or rain), and providing velocity estimation from a single measurement.
However, radar also brings its challenges such as sparsity (typically $180\times$ fewer than LiDAR points per single frame in nuScenes~\cite{Caesar2020}), noisy and ambiguous measurements (false negatives by low resolution, accuracy, or low radar cross-section, and false positives by multi-path or clutters).
As a result, previous camera-radar fusion methods using late fusion strategies that fuse detection-level results~\cite{cho2014multi, gohring2011radar} fail to fully exploit the complementary information, thus having limited performance and operating environment.
Despite the huge potential of learning-based fusion, only a few studies~\cite{kim2020grif, kim2022craft, nabati2021centerfusion} explore camera-radar fusion in autonomous driving scenarios.

To put the aforementioned advantages and disadvantages of camera and radar in perspective, camera-radar fusion should be capable of following properties to fully exploit the complementary characteristics of each sensor.
First, camera features should be accurately transformed into BEV space in terms of spatial position.
Second, the fusion method should be able to handle the spatial misalignment between feature maps when aggregating two modalities.
Last but not least, transformation and fusion should be adaptive in order to tackle noisy and ambiguous radar measurements.

To this end, we design a novel two-stage fusion method for BEV feature encoding, \textit{Camera Radar Net (CRN)}. 
The key idea of the proposed method is to generate \textit{semantically rich and spatially accurate BEV feature map} by fusing complementary characteristics of camera and radar sensors.
In particular, we first transform image features in perspective view into BEV not solely relying on estimated depth but using radar, named \textit{radar-assisted view transformation (RVT)}.
Since transformed image features in BEV is not completely accurate, following \textit{multi-modal feature aggregation (MFA)} layers consecutively encodes the multi-modal feature maps into a unified feature map using an attention mechanism.
We conduct extensive experiments on nuScenes and demonstrate that our proposed method can generate a fine-grained BEV feature map to set the new state-of-the-art on various tasks while maintaining high efficiency, as shown in Fig~\ref{fig:intro}.

The main contributions of our works are three-fold:
\begin{itemize}
\setlength\itemsep{-3pt}
\item \textbf{Accuracy.}\hspace{0.2cm} CRN achieves LiDAR-level performance using camera and radar on 3D object detection, tracking, and BEV segmentation tasks.
\item \textbf{Robustness.}\hspace{0.2cm} CRN maintains robust performance even if one of the single sensor inputs is entirely unavailable, which allows the fault-tolerant system.
\item \textbf{Efficiency.}\hspace{0.2cm} CRN requires marginal extra cost for significant performance improvement, which enables long-range perception in real-time.
\end{itemize}

\section{Related Work}\label{sec:related_works}

\noindent
\textbf{Camera-based 3D Perception.\hspace{0.2cm}}
Thanks to well-established 2D object detection methods~\cite{tian2019fcos, Zhou2019} on perspective view images, early approaches extend 2D detector to 3D detector by additionally estimating the distance to objects~\cite{Simonelli, wang2021fcos3d}, then transforming object center.
DD3D~\cite{park2021pseudo} improves detection performance by pre-training depth estimation task on depth dataset~\cite{Guizilini2020}.
Although a simple and intuitive approach, the view discrepancy between input feature space (perspective view, PV) and output space (bird's-eye-view, BEV) restricts the network from extending to other tasks.

Recent advances in camera-based perception exploit view transformation.
Geometry-based methods~\cite{li2022bevdepth, park2022time, philion2020lift, reading2021categorical} explicitly estimate the depth distribution of each image feature on PV and transform them into BEV.
BEVDepth~\cite{li2022bevdepth} empirically shows that training depth distribution with auxiliary pixel-wise depth supervision improves the performance, which corresponds to the results of DD3D~\cite{park2021pseudo}.
Learning-based methods~\cite{jiang2022polarformer, li2022bevformer, lu2022learning, zhou2022cross} implicitly model the mapping function from PV to BEV using multi-layer perceptron (MLP)~\cite{li2022hdmapnet, roddick2020predicting} or attention~\cite{li2022bevformer, saha2022translating}.

Obtaining a BEV feature map allows the framework to be easily extended to various downstream tasks performed on BEV space, such as 3D detection and tracking~\cite{li2022bevdepth}, segmentation~\cite{zhou2022cross}, and prediction~\cite{philion2020lift}.
However, camera-only methods have limited localization accuracy due to the absence of distance information in image and are sensitive to lighting or weather conditions.
Moreover, achieving high performance only using a camera requires large image input and backbone, which is slow and not applicable for real-time applications.

\vspace{5pt}
\noindent\textbf{Point-based 3D Perception.\hspace{0.2cm}}
LiDAR is the most common and favorable sensor for autonomous driving, while radar point cloud has not yet been thoroughly investigated.
LiDAR-only 3D detectors extract features (\textit{e.g.}, PointNet~\cite{qi2017pointnet, qi2017pointnet++}) given irregular and unordered point sets and predict 3D objects on point-~\cite{shi2019pointrcnn} or voxelized-~\cite{lang2019pointpillars} feature.
Some approaches further utilize point and voxel features together~\cite{Shi2020a}, use range view as additional features~\cite{wang2020pillar}, or filter background points~\cite{sun2021rsn}.

Although similar data representation of radar point cloud~\cite{Caesar2020, meyer2019automotive} to LiDAR, radar point-based 3D perception is considerably less investigated.
Several works~\cite{popov2022nvradarnet, sless2019road, weston2019probably} examine the radar points for free space detection, but only a few studies~\cite{svenningsson2021radar, ulrich2022improved} attempt 3D object detection in autonomous driving.
Radar point-based detection methods adapt PointPillars~\cite{lang2019pointpillars} with graph neural network~\cite{shi2020point} or KPConv~\cite{thomas2019kpconv} focusing on extracting better local features.
However, mostly due to many clutter points and lack of contextual information on radar, the performance of radar-only methods lags significantly behind compared to LiDAR.
Considering the high potential of radar having robust measurements regardless of weather conditions and perception range, fusing radar with a camera is promising to supplement the insufficient semantic information.

\begin{figure*}[t]
\begin{center}
\includegraphics[width=0.96\textwidth]{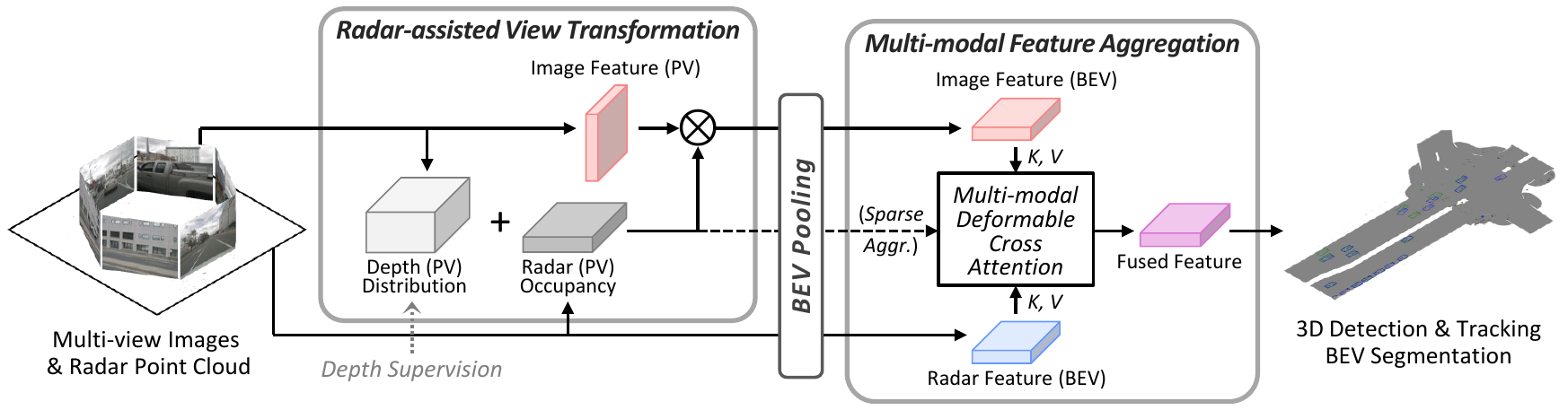}
\end{center}
\vspace{-8pt}
\caption{
    The overall architecture of the proposed Camera Radar Net.
    Given multi-view images and radar points, modality-specific backbones extract features in each view.
    First, image context features in perspective view are transformed into a bird's-eye-view with the help of radar measurements by Radar-assisted View Transformation (RVT).
    After, Multi-modal Feature Aggregation (MFA) adaptively aggregates image and radar feature maps to generate semantically rich and spatially accurate bird's-eye-view representation.
}
\label{fig:architecture}
\end{figure*}

\vspace{5pt}
\noindent\textbf{Camera-Point 3D Perception.\hspace{0.2cm}}
Fusing complementary information of camera image and range measurement is a promising and active research topic.
However, the view discrepancy between two sensors is regarded as a bottleneck for multi-modal fusion.
A line of approach handles discrepancy by projecting 3D information to a 2D image (\textit{e.g.}, points~\cite{chen2022autoalignv2, vora2020pointpainting, wu2023mvfusion}, proposals~\cite{bai2022transfusion, kim2020grif, ku2018joint}, or prediction results~\cite{pang2020clocs}) and gathering information around the projected region.
Some camera-radar fusion methods~\cite{lin2020depth, long2021radar} attempt to improve depth estimation by projecting radar points to the image.

On the other hand, another line of work lifts 2D image information into 3D.
Early studies in 3D detection~\cite{kim2022craft, nabati2021centerfusion, qi2018frustum} detect 2D or 2.5D object proposals and lift them into 3D space to fuse with point data; however, this object-level fusion is difficult to be generalized to other tasks in BEV.
Thanks to advances in monocular BEV approaches, recent fusion approaches extract image and point feature maps in unified BEV space and then fuse feature maps by element-wise concatenation~\cite{liu2022bevfusion} or summation~\cite{li2022unifying}, assuming multi-modal feature maps are spatially well aligned.
After, the fused BEV feature map is used in various perception tasks such as 3D detection~\cite{drews2022deepfusion, li2022unifying, liang2022bevfusion, yoo20203d}, BEV segmentation~\cite{liu2022bevfusion, zhou2023bridging}, or HD map generation~\cite{dong2022superfusion, li2022hdmapnet}.
However, despite the unique characteristics of a camera (\textit{e.g.}, inaccurate BEV transformation) and radar (\textit{e.g.}, sparsity and ambiguity), previous camera-radar fusion less considers them.
Our proposed CRN focuses on fusing multi-modal feature maps considering the characteristics of each sensor thoroughly to have the best of both worlds.

\section{Camera Radar Net} \label{sec:method}
In this paper, we propose a camera radar fusion framework to produce a unified BEV representation given multi-view images and radar points, as illustrated in Fig~\ref{fig:architecture}.
In Sec.~\ref{sec:RVT}, we introduce a method to transform image features with radar, then a multi-modal feature aggregation method in Sec.~\ref{sec:MFA}.
Finally, generated BEV feature map is used for downstream tasks in Sec.~\ref{sec:heads}.

\subsection{Preliminaries} \label{sec:preliminaries}
\noindent\textbf{Monocular 3D Approaches.\hspace{0.2cm}}
The crux of monocular 3D perception is \textit{how to construct accurate 3D (or BEV) information from 2D features}, which can be categorized into two groups.
Geometry-based approaches~\cite{li2022bevdepth, philion2020lift, reading2021categorical} predict depth $\mathbf{D}$ as an explicit intermediate representation and transform features $\mathbf{F}$ in perspective view ($u,v$) into frustum view ($d,u,v$) then 3D ($x,y,z$) by:
\begin{equation}
    \mathbf{F}_{3D}(x,y,z)=\mathcal{M}(\mathbf{F}_{2D}(u,v) \otimes \mathbf{D}(u,v)),
\label{eq:LSS}
\end{equation}
where $\mathcal{M}$ denotes view transformation module (\textit{e.g.}, Voxel Pooling~\cite{li2022bevdepth, liu2022bevfusion}) and $\otimes$ denotes outer product.
Meanwhile, learning-based approaches~\cite{li2022bevformer, zhou2022cross} implicitly model 3D to 2D projection utilizing mapping networks as:
\begin{equation}
    \mathbf{F}_{3D}(x,y,z)=f(P_{xyz}, \mathbf{F}_{2D}(u,v)),
\end{equation}
where $f$ denotes mapping function between perspective view and BEV (\textit{e.g.}, multi-layer perceptron (MLP)~\cite{roddick2020predicting} or cross-attention~\cite{li2022bevformer}), and $P_{xyz}$ is voxels in 3D space.
Although the approaches are different, the key is to obtain spatially accurate 3D features $\mathbf{F}_{3D}(x,y,z)$ through implicit or explicit transformation.
We aim to explicitly improve the transformation process using radar measurement.

\vspace{3pt}
\noindent\textbf{Radar Characteristics.\hspace{0.2cm}}
Radar can have various representations (\textit{e.g.}, 2-D FFT~\cite{lin2018human}, 3D Tensor~\cite{kim2020low, major2019vehicle}, point cloud~\cite{Caesar2020, meyer2019automotive}).
Radar point cloud has a similar representation to LiDAR, but their characteristics are different in terms of resolution and accuracy~\cite{ars408}.
Moreover, due to the nature of the operating mechanism of radar~\cite{johnson1992array, li2008mimo} and its millimeter scale wavelength, radar measurements are noisy, ambiguous, and do not provide elevation.
Therefore, radar measurements are often not returned when objects exist or returned when objects do not exist; hence, naively adopting LiDAR methods to radar shows very limited performance on complex scenarios, as in Tables~\ref{table:ablation_range} and \ref{table:analysis_robust} (CenterPoint~\cite{yin2021center} with radar input).
We exploit radar in an adaptive manner to handle its sparsity and ambiguity.


\begin{figure}[t]
\begin{center}
\includegraphics[width=0.94\columnwidth]{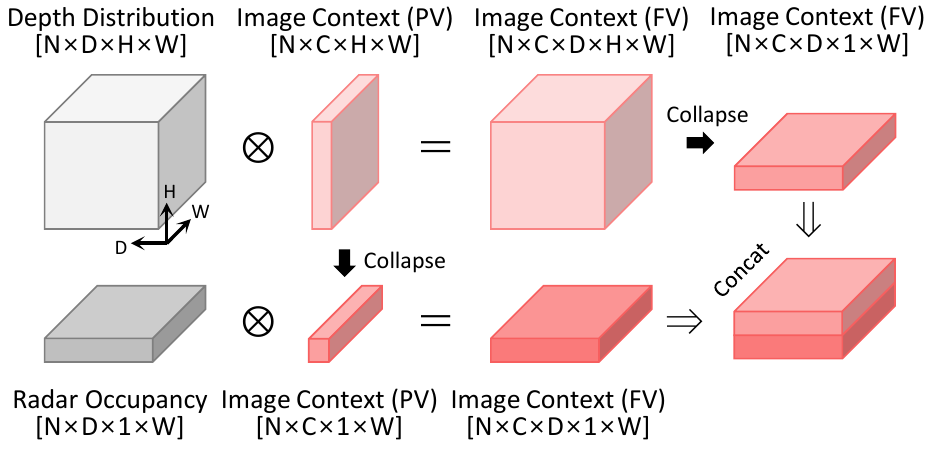}
\end{center}
\vspace{-10pt}
\caption{
    Radar-assisted View Transformation (RVT).
    The proposed RVT can benefit from dense but less accurate depth distribution and sparse but accurate radar occupancy to obtain spatially accurate image context features.
}
\label{fig:RVT}
\end{figure}

\subsection{Radar-assisted View Transformation (RVT)} \label{sec:RVT}
\noindent\textbf{Image Feature Encoding and Depth Distribution.\hspace{0.1cm}}
Given a set of $N$ surrounding images, we use an image backbone (\textit{e.g.}, ResNet~\cite{he2016deep}, ConvNeXt~\cite{liu2022convnet}) with a feature pyramid network (FPN)~\cite{lin2017feature} and obtain $16\times$ downsampled feature map $\mathbf{F}_{I}$ for each image view.
Then, additional convolutional layers further extract image context features $\mathbf{C}_{I}^{PV}\in {\mathbb{R}}^{N\times C\times H\times W}$ and depth distribution of each pixel $\mathbf{D}_{I}\in {\mathbb{R}}^{N\times D\times H\times W}$ in perspective view, following LSS~\cite{philion2020lift}:

\begin{equation}
\begin{split}
    & \mathbf{C}_{I}^{PV}=\text{Conv}(\mathbf{F}_{I}) \\ 
    & \mathbf{D}_{I}(u,v)=\text{Softmax}(\text{Conv}(\mathbf{F}_{I})(u,v)),
\label{eq:image feature extract}
\end{split}
\end{equation}
where $(u,v)$ indicates coordinate in the image plane, and $D$ is the number of depth bins.

\vspace{3pt}
\noindent\textbf{Radar Feature Encoding and Radar Occupancy.\hspace{0.2cm}}
Unlike previous methods~\cite{li2022bevdepth, philion2020lift, reading2021categorical} that directly ``lift'' image features into BEV using \textit{estimated} depth distribution as Eq.~\ref{eq:LSS}, we exploit noisy yet accurate radar measurements for view transformation.
Radar points are first projected onto each $N$ camera view to find corresponding image pixels while preserving its depth, then voxelized~\cite{lang2019pointpillars} into camera frustum view voxels $\mathbf{V}_{R}^{FV}(d,u,v)$.
Note that $u,v$ are pixel units in the image width and height directions, while $d$ is a metric unit in a depth direction.
We set $v=1$ to use pillar-style since radars do not provide reliable elevation measurements.
The non-empty radar pillars are encoded into features $\mathbf{F}_{R}\in {\mathbb{R}}^{N\times C\times D\times W}$ with PointNet~\cite{qi2017pointnet++} and sparse convolution~\cite{yan2018second}.
Similar to Eq.~\ref{eq:image feature extract}, we extract radar context feature $\mathbf{C}_{R}^{FV}\in {\mathbb{R}}^{N\times C\times D\times W}$ and radar occupancy $\mathbf{O}_{R}\in {\mathbb{R}}^{N\times 1\times D\times W}$ in frustum view. 
Here, convolution is applied to top-view $(d,u)$ coordinate instead of $(u,v)$:
\begin{equation}
    \mathbf{C}_{R}^{FV}=\text{Conv}(\mathbf{F}_{R}),\; \mathbf{O}_{R}(d,u)=\sigma(\text{Conv}(\mathbf{F}_{R})(d,u)).
\label{eq:radar feature extract}
\end{equation}
Here, a sigmoid is used instead of softmax since radar occupancy is not necessarily one-hot encoded as a depth distribution.

\vspace{3pt}
\noindent\textbf{Frustum View Transformation.\hspace{0.2cm}}
Given depth distribution $\mathbf{D}_{I}$ and radar occupancy $\mathbf{O}_{R}$, the image context feature map $\mathbf{C}_{I}^{PV}$ is transformed into a camera frustum view $\mathbf{C}_{I}^{FV}\in {\mathbb{R}}^{N\times C\times D\times H\times W}$ as:
\begin{equation}
    \mathbf{C}_{I}^{FV}=\text{Conv}[\mathbf{C}_{I}^{PV} \otimes \mathbf{D}_{I}; \mathbf{C}_{I}^{PV} \otimes \mathbf{O}_{R}],
\end{equation}
where $[\cdot;\cdot]$ denotes the concatenation operating along the channel dimension and $\otimes$ is the outer product.
Due to the absence of height dimension in radar and for saving memory, we collapse the image context feature by summation along the height axis, as illustrated in Fig.~\ref{fig:RVT}.

\vspace{3pt}
\noindent\textbf{Bird's-Eye-View Transformation.\hspace{0.2cm}}
Finally, camera and radar context feature maps in $N$ camera frustum views $\mathbf{F}^{FV}=\{\mathbf{C}_{I}^{FV}, \mathbf{C}_{R}^{FV} \in \mathbb{R}^{N\times C\times D\times H\times W}\}$ are transformed into a single BEV space $\mathbb{R}^{C\times 1\times X\times Y}$ by view transformation module $\mathcal{M}$:
\begin{equation}
    \mathbf{F}^{BEV}=\mathcal{M}(\{\mathbf{F}_{i}^{FV}\}_{i=1}^{N}).
\label{eq:BEV feature}
\end{equation}
Specifically, we adopt CUDA-enabled Voxel Pooling~\cite{li2022bevstereo} implementation and modify it to aggregate features within each BEV grid using average pooling instead of summation.
It helps the network to predict a more consistent BEV feature map regardless of the distance to the ego vehicle since a closer BEV grid is associated with a more frustum grid due to the perspective projection.

\subsection{Multi-modal Feature Aggregation (MFA)} \label{sec:MFA} 
\noindent\textbf{Motivation.\hspace{0.2cm}}
Combining complementary multi-modal information while avoiding the drawbacks of each is especially crucial in camera radar fusion, as claimed in Sec.~\ref{sec:preliminaries}.
Image feature has rich semantic cues, but their spatial position is inherently inaccurate; on the other hand, radar feature is spatially accurate, but contextual information is insufficient and noisy.
Naive approaches are channel-wise concatenation~\cite{liu2022bevfusion} or summation~\cite{li2022unifying}, but these cannot handle neither spatial misalignment nor ambiguity between two modalities, thus less effective, as can be seen in Table~\ref{table:ablation_MFA}.
To have the best of both worlds, the key motivation of our fusion is to leverage multi-modal features in an adaptive manner, using an attention mechanism~\cite{vaswani2017attention}.

\vspace{3pt}
\noindent\textbf{Multi-modal Deformable Cross Attention (MDCA).\hspace{0.2cm}}
Cross attention~\cite{vaswani2017attention} is inherently suitable for multi-modal fusion.
However, the computation cost is quadratic to input sequence length $\mathcal{O}(N^2)$, where $N=XY$ and $X, Y$ denote the height and width of the BEV feature map.
If we assume perception range $R=X/2=Y/2$, computation complexity becomes biquadratic $\mathcal{O}(16R^4)$ to perception range, which is not scalable for a long-range perception;
Thus we develop the fusion method based on deformable attention~\cite{zhu2020deformable}, which is of linear complexity with the input size $\mathcal{O}(2N+NK)$, where $K$ is the total number of the sampled key ($K\ll N=XY$).

\begin{figure}[t]
\begin{center}
\includegraphics[width=0.98\columnwidth]{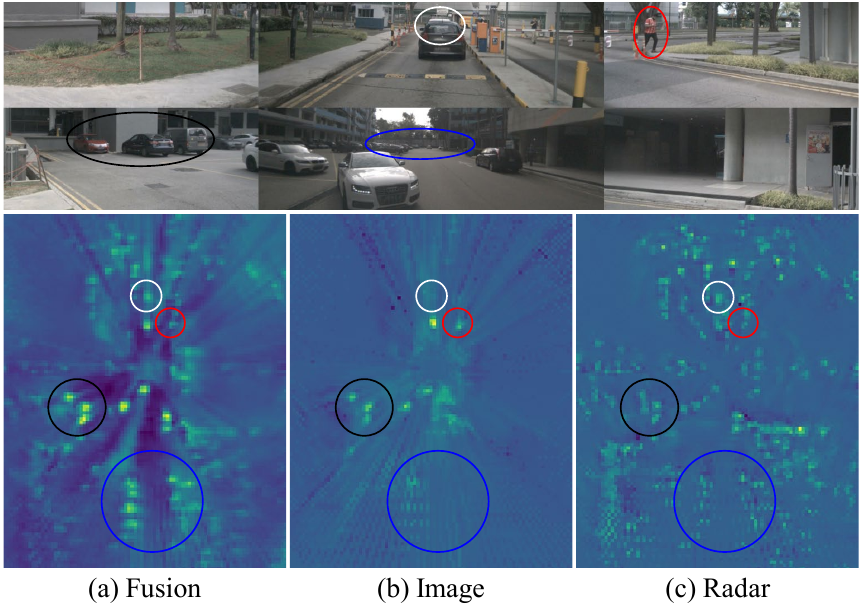}
\end{center}
\vspace{-6pt}
\caption{
    Visualization of feature maps trained on detection task.
    In image, a vehicle heavily occluded (white) or hardly visible at a long distance (blue) is not detected.
    In radar, clutters from the wall (black) or pedestrian with row RCS (red) lead to failure.
    Our MFA generates a more reliable BEV feature map by fusion.
    Note that BEV feature maps are cropped for better visualization.
}
\label{fig:feature visualize}
\vspace{-8pt}
\end{figure}

Given flattened BEV context feature maps $\mathrm{x}_m=\{\mathbf{C}_{I}^{BEV}, \mathbf{C}_{R}^{BEV} \in \mathbb{R}^{C\times XY}\}$, we first project $\mathrm{x}_m$ into $C$ dimensional query feature after concatenation as $\mathrm{z}_q=\boldsymbol{W}_z[\text{LN}(\mathbf{C}_{I});\text{LN}(\mathbf{C}_{P})]$, where $\boldsymbol{W}_z \in\mathbb{R}^{C\times 2C}$ is a linear projection and $\text{LN}$ is layer norm.
After, the feature map is aggregated by multi-modal deformable cross attention as
\begin{equation}
\begin{split}
    & \text{MDCA}(\mathrm{z}_q,p_q,\mathrm{x}_m)= \\
    & \sum_{h}^{H} \boldsymbol{W}_h \left[ \sum_{m}^{M} \sum_{k}^{K} A_{hmqk} \cdot \boldsymbol{W}'_{hm} \mathrm{x}_m (\phi_{m}(p_q + \Delta p_{hmqk})) \right],
\end{split}
\end{equation}
where $h, m, k$ indexes the attention head, modality, and sampling point.
To better exploit multi-modal information, we separately apply attention weights $A_{hmqk}$ and sampling offset $\Delta p_{hmqk}$ to multi-modal feature maps $\mathrm{x}_m$.
By doing so, the feature aggregation module can adaptively benefit from image and radar as shown in Fig.~\ref{fig:feature visualize}.
We refer the reader to Appendix for details of the notation. 

\vspace{3pt}
\noindent\textbf{Sparse Aggregation.\hspace{0.2cm}}
Although MDCA has linear complexity with respect to the size of BEV grids, it still can be a bottleneck when the perception range becomes large.
Inspired by~\cite{roh2021sparse}, we propose a method to further reduce the number of input queries from $N=XY$ to $N=N_k \ll XY$ by using features with top-k confidence.
Given BEV depth distribution $\mathbf{D}_{I}$ and radar occupancy $\mathbf{O}_{P}$, $N_k$ features $\mathrm{z}^{N_k}_q \in {\mathbb{R}}^{C \times N_k}$ are selected from input queries $\mathrm{z}_q \in {\mathbb{R}}^{C \times XY}$ using a probability of $\max(\mathbf{D}_{I}, \mathbf{O}_{P})$.
The complexity of the proposed sparse aggregation is now independent of perception range, which is more efficient for long-range perception.

\subsection{Training Objectives and Task Heads} \label{sec:heads}
For all tasks, we train the depth distribution network with a depth map obtained by projecting LiDAR points into the image view, following BEVDepth~\cite{li2022bevdepth}.

\vspace{3pt}
\noindent\textbf{3D Detection and Tracking.\hspace{0.2cm}}
For the 3D object detection task, we follow CenterPoint~\cite{yin2021center} to predict the center heatmap with anchor-free and multi-group head~\cite{zhu2019class}.
After, we perform 3D tracking by tracking-by-detection using velocity-based closest distance matching~\cite{yin2021center}.
For training sparse aggregation setting, we filter LiDAR points outside of the 3D bounding box when obtaining a ground truth depth map and replace the softmax to sigmoid in Eq.~\ref{eq:image feature extract}; thereby, only feature grids containing foreground objects can have a high probability.

\vspace{3pt}
\noindent\textbf{BEV Segmentation.\hspace{0.2cm}}
For the BEV segmentation task, we attach a convolutional decoder head to obtain the prediction map following CVT~\cite{zhou2022cross}.
Given a BEV feature map from Multi-modal Feature Aggregation (MFA) layers, the segmentation head encodes to a latent representation and decodes back to the final output segmentation map, followed by a sigmoid layer.
Our BEV segmentation network predicts a semantic occupancy grid of vehicles and drivable area, trained with a focal loss~\cite{Lin2017a}.


\section{Experiments}\label{sec:experiments}

\setlength{\tabcolsep}{0.65em}
\begin{table*}[!t]
\begin{center}
\resizebox{0.98\textwidth}{!}{
\begin{tabular}{l|c|c|c||cc|ccccc|c}
    \hline
    Method & Input & Backbone & Image Size & NDS$\uparrow$ & mAP$\uparrow$ & mATE$\downarrow$ & mASE$\downarrow$ & mAOE$\downarrow$ & mAVE$\downarrow$ & mAAE$\downarrow$ & FPS \\
    \hline
    CenterPoint-P$^{\dagger*}$ \cite{yin2021center} 
    & L & Pillars & - & 59.8 & 49.4 & 0.320 & 0.262 & 0.377 & 0.334 & 0.198 & - \\
    CenterPoint-V$^{\dagger*}$ \cite{yin2021center} 
    & L & Voxel   & - & 65.3 & 56.9 & 0.285 & 0.253 & 0.323 & 0.272 & 0.186 & - \\
    \hline
    BEVDet$^{\dagger}$ \cite{huang2021bevdet}
    & C & R50 & $256\times704$ & 39.2 & 31.2 & 0.691 & 0.272 & 0.523 & 0.909 & 0.247 & 15.6 \\
    CenterFusion$^{\dagger}$ \cite{nabati2021centerfusion}
    & C+R & DLA34 & $448\times800$ & 45.3 & 33.2 & 0.649 & \textbf{0.263} & 0.535 & 0.540 & \textbf{0.142} & - \\    
    BEVDepth$^{\dagger}$ \cite{li2022bevdepth}
    & C & R50 & $256\times704$ & 47.5 & 35.1 & 0.639 & 0.267 & 0.479 & 0.428 & 0.198 & 11.6 \\
    RCBEV4d$^{\dagger}$ \cite{zhou2023bridging}
    & C+R & Swin-T & $256\times704$ & 49.7 & 38.1 & 0.526 & 0.272 & 0.445 & 0.465 & 0.185 & - \\
    CRAFT$^{\dagger}$ \cite{kim2022craft}
    & C+R & DLA34 & $448\times800$ & 51.7 & 41.1 & 0.494 & 0.276 & 0.454 & 0.486 & 0.176 & 4.1 \\
    SOLOFusion$^{\dagger}$ \cite{park2022time}
    & C & R50 & $256\times704$ & 53.4 & 42.7 & 0.567 & 0.274 & \textbf{0.411} & \textbf{0.252} & 0.188 & 11.4 \\
    \gr \textbf{CRN}
    & C+R & R18 & $256\times704$ & 54.3 & 44.8 & 0.518 & 0.283 & 0.552 & 0.279 & 0.180 & \textbf{27.9} \\
    \gr \textbf{CRN}
    & C+R & R50 & $256\times704$ & \textbf{56.0} & \textbf{49.0} & \textbf{0.487} & 0.277 & 0.542 & 0.344 & 0.197 & 20.4 \\
    \hline
    PETR$^{\dagger}$ \cite{liu2022petr}
    & C & R101 & $900\times1600$ & 44.2 & 37.0 & 0.711 & 0.267 & 0.383 & 0.865 & 0.201 & 1.7 \\
    MVFusion$^{\dagger}$ \cite{wu2023mvfusion}
    &C+R& R101 & $900\times1600$ & 45.5 & 38.0 & 0.675 & 0.258 & 0.372 & 0.833 & 0.196 & - \\
    BEVFormer \cite{li2022bevformer}
    & C & R101 & $900\times1600$ & 51.7 & 41.6 & 0.673 & 0.274 & 0.372 & 0.394 & 0.198 & 1.7 \\
    BEVDepth$^{\dagger}$ \cite{li2022bevdepth}
    & C & R101 & $512\times1408$ & 53.5 & 41.2 & 0.565 & 0.266 & \textbf{0.358} & 0.331 & 0.190 & 5.0 \\
    SOLOFusion \cite{park2022time}
    & C & R101 & $512\times1408$ & 54.4 & 47.2 & 0.518 & 0.275 & 0.604 & 0.310 & 0.210 & - \\
    SOLOFusion$^{\dagger}$ \cite{park2022time}
    & C & R101 & $512\times1408$ & 58.2 & 48.3 & 0.503 & 0.264 & 0.381 & \textbf{0.246} & 0.207 & - \\
    \gr \textbf{CRN}
    & C+R & R101 & $512\times1408$ & 59.2 & 52.5 & 0.460 & 0.273 & 0.443 & 0.352 & 0.180 & \textbf{7.2} \\
    \gr \textbf{CRN}$^{\ddagger}$
    & C+R & R101 & $512\times1408$ & \textbf{60.7} & \textbf{54.5} & \textbf{0.445} & 0.268 & 0.425 & 0.332 & 0.180 & - \\
    \hline
\end{tabular}}
\end{center}
\vspace{-5pt}
\caption{
    \textbf{3D Object Detection} on nuScenes \texttt{val} set. 
    `L', `C', and `R' represent LiDAR, camera, and radar, respectively.
    $^*$: results from MMDetection3D~\cite{chen2019mmdetection}.
    $^\dagger$: trained with CBGS.
    $^\ddagger$: use test time augmentation.
}
\vspace{-8pt}
\label{table:3D det val set}
\end{table*}

\renewcommand{\thefootnote}{1}
\setlength{\tabcolsep}{0.4em}
\begin{table}[!t]
\begin{center}
\resizebox{1.0\columnwidth}{!}{
\begin{tabular}{l|c|c||cc|c}
    \hline
    Method & Input  & Backbone & NDS$\uparrow$ & mAP$\uparrow$ & mATE$\downarrow$\\
    \hline
    PointPillars \cite{lang2019pointpillars}    & L & Pillars  & 55.0 & 40.1 & 0.392 \\
    CenterPoint \cite{yin2021center}            & L & Voxel    & 67.3 & 60.3 & 0.262 \\
    \hline
    \small{KPConvPillars} \cite{ulrich2022improved}   & R & Pillars    & 13.9 &  4.9 & 0.823 \\
    \hline
    CenterFusion \cite{nabati2021centerfusion}  &C+R& DLA34    & 44.9 & 32.6 & 0.631 \\
    RCBEV \cite{zhou2023bridging}               &C+R& Swin-T   & 48.6 & 40.6 & 0.484 \\
    PETR \cite{liu2022petr}                     & C & V2-99    & 50.4 & 44.1 & 0.593 \\
    MVFusion \cite{wu2023mvfusion}              &C+R& V2-99    & 51.7 & 45.3 & 0.569 \\
    CRAFT \cite{kim2022craft}                   &C+R& DLA34    & 52.3 & 41.1 & 0.467 \\
    BEVFormer \cite{li2022bevformer}            & C & V2-99    & 56.9 & 48.1 & 0.582 \\
    BEVDepth \cite{li2022bevdepth}              & C & \small{ConvNeXt-B} & 60.9 & 52.0 & 0.445 \\
    BEVStereo \cite{li2022bevstereo}            & C & V2-99    & 61.0 & 52.5 & 0.431 \\
    SOLOFusion \cite{park2022time}              & C & \small{ConvNeXt-B} & 61.9 & 54.0 & 0.453 \\
    \gr \textbf{CRN}                            &C+R& \small{ConvNeXt-B} & \textbf{62.4} & \textbf{57.5} & \textbf{0.416} \\
    \hline
\end{tabular}}
\end{center}
\vspace{-5pt}
\caption{
    \textbf{3D Object Detection} on nuScenes \texttt{test} set.
    V2-99 is pre-trained on external depth dataset DDAD~\cite{Guizilini2020}.
}
\vspace{-8pt}
\label{table:3D det test set}
\end{table}

\subsection{Experimental Settings}
\noindent\textbf{Dataset and Metrics.\hspace{0.2cm}}
We conduct experiments on nuScenes~\cite{Caesar2020}, which provides radar point clouds at scale.
For 3D object detection and tracking, we use official metrics: mAP~\cite{everingham2010pascal}, NDS~\cite{Caesar2020}, and AMOTA~\cite{weng2019baseline}, and we follow the settings proposed by LSS~\cite{philion2020lift} for BEV segmentation.
We refer the reader to nuScenes~\cite{Caesar2020} and LSS~\cite{philion2020lift} for details of metrics.

\vspace{3pt}
\noindent\textbf{Implementation Details.\hspace{0.2cm}}
For the camera stream, we adopt BEVDepth~\cite{li2022bevdepth} as a baseline with several modifications.
We reduce the number of depth estimation layers and eliminate the depth refinement module, which increases the inference speed without a significant performance drop.
For radar, we accumulate six previous radar sweeps and use normalized RCS and Doppler speed as features following GRIF Net~\cite{kim2020grif}.
Unless otherwise specified, we follow standard practices~\cite{li2022bevdepth} for implementation and training details.
We accumulate previous three BEV feature maps with an interval of 1 second, similar to BEVFormer~\cite{li2022bevformer}.

Our models are trained for 24 epochs with AdamW~\cite{loshchilov2018decoupled} optimizer in an end-to-end manner, unless otherwise specified.
In addition to image and BEV data augmentation~\cite{li2022bevdepth}, we randomly drop sweeps and points for radar~\cite{leng2022lidaraugment}.
Inference time is measured on an Intel Core i9 CPU and RTX 3090 GPU with a single batch and FP16 precision.
The full experimental settings are provided in Appendix.

\subsection{Main Results}
\begin{figure*}[!t]
\begin{center}
\includegraphics[width=0.96\textwidth]{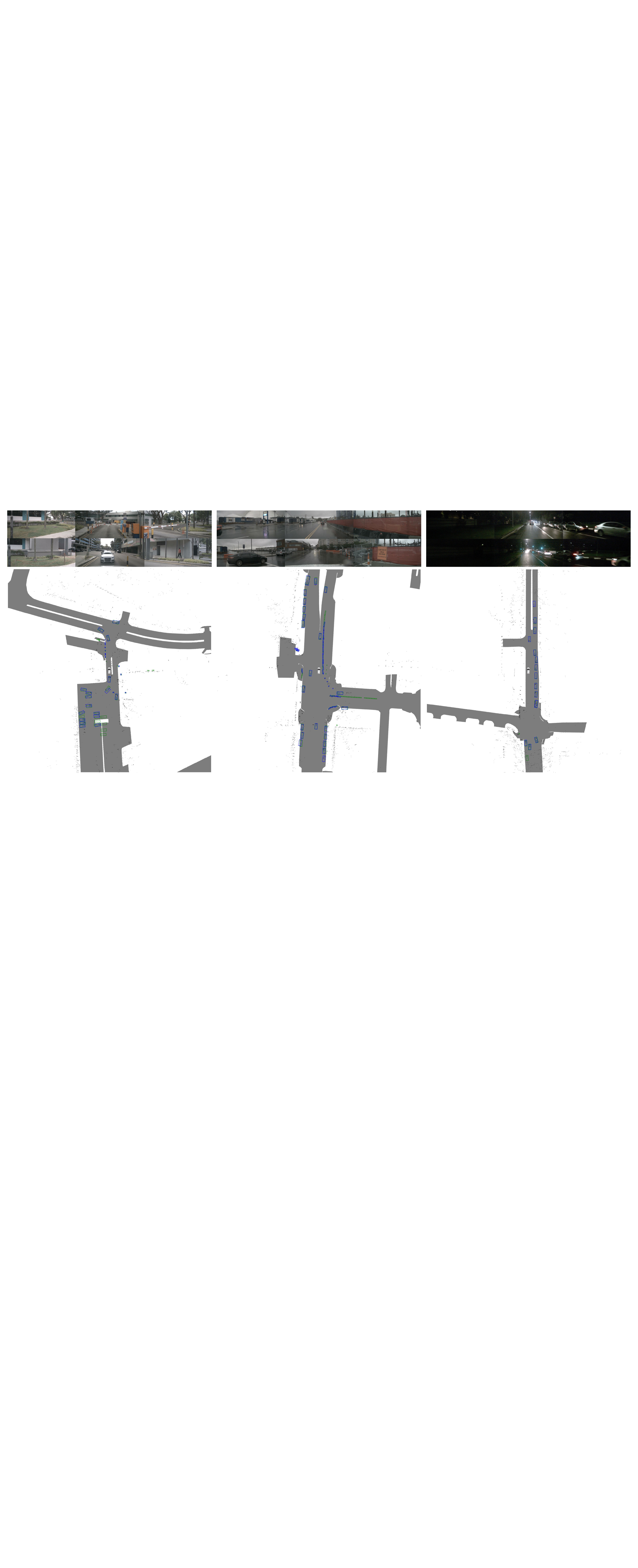}
\end{center}
\vspace{-8pt}
\caption{
Qualitative results of long-range model on various weather conditions.
Images on the top are the six camera views surrounding the vehicle.
Green boxes are ground truths, blue boxes are our prediction results, and black dots are radar points.
Perception ranges are set to $200m\times200m$, and ground truth maps on the background are used for visualization. 
Best viewed in color with zoom in.
}
\label{fig:results}
\vspace{-10pt}
\end{figure*}

\noindent\textbf{3D Object Detection.\hspace{0.2cm}}
For a fair comparison with previous state-of-the-art 3D detection methods, we train our model only on 3D detection task and report \texttt{val} and \texttt{test} set results in Tables~\ref{table:3D det val set} and \ref{table:3D det test set}.
Under various input image sizes and backbone settings, our CRN ranks first place among all camera-only and camera-radar methods with much faster FPS (Sec. \ref{sec:inference time} for inference time analysis).
We emphasize that CRN with a small image input and backbone ($256\times 704$ and R18) already outperforms competitors with a large image input and backbone (BEVFormer~\cite{li2022bevformer} and BEVDepth~\cite{li2022bevdepth} with $512\times 1408$ and R101) in terms of mAP while running an order of magnitude faster, showing the effectiveness of using radar over camera-only methods.
CRN also outperforms the LiDAR method CenterPoint-P~\cite{yin2021center}, demonstrating the potential of cost-effective camera and radar to replace LiDAR for autonomous driving.
Qualitative results are provided in Fig.~\ref{fig:results} and Appendix.

\renewcommand{\thefootnote}{2}
\setlength{\tabcolsep}{0.3em}
\begin{table}[!t]
\begin{center}
\resizebox{1.0\columnwidth}{!}{
\begin{tabular}{l|c||cc|ccc}
    \hline
    Method & Input  & AMOTA$\uparrow$ & AMOTP$\downarrow$ & FP$\downarrow$ & FN$\downarrow$ & IDS$\downarrow$\\
    \hline
    CenterPoint \cite{yin2021center}    & L & 63.8 & 0.555 & 18612 & 22928 & 760 \\
    \hline
    DEFT \cite{chaabane2021deft}        & C & 17.7 & 1.564 & 22163 & 60565 & 6901 \\
    QD-3DT \cite{hu2022monocular}       & C & 21.7 & 1.550 & \textbf{16495} & 60156 & 6856 \\
    CC-3DT \cite{fischer2022cc}         & C & 41.0 & 1.274 & 18114 & 42910 & 3334 \\
    Sparse4D \cite{lin2022sparse4d}     & C & 51.9 & 1.078 & 19626 & \textbf{32954} & 1090 \\
    \gr \textbf{CRN}                    &C+R& \textbf{56.9} & \textbf{0.809} & 16822 & 41093 & \textbf{946} \\ 
    \hline
\end{tabular}}
\end{center}
\vspace{-5pt}
\caption{
    \textbf{3D Object Tracking} on nuScenes \texttt{test} set.
}
\vspace{-8pt}
\label{table:3D Track test set}
\end{table}

\setlength{\tabcolsep}{0.3em}
\begin{table}[!t]
\begin{center}
\resizebox{1.0\columnwidth}{!}{
\begin{tabular}{l|c|c|c||cc|c}
    \hline
    Method & Input & Backbone & Image Size & Veh. & D.A. & FPS\\
    \hline
    LSS \cite{philion2020lift}          & C & \small{EffNetB0} & $128\times352$ & 32.1 & 72.0 & 25 \\
    FIERY \cite{hu2021fiery}            & C & \small{EffNetB4} & $224\times480$ & 35.8 & - & 8 \\
    CVT \cite{zhou2022cross}            & C & \small{EffNetB4} & $224\times448$ & 36.0 & 74.3 & 35 \\
    GKT \cite{chen2022efficient}        & C & \small{EffNetB4} & $224\times448$ & 38.0 & - & \textbf{45.6}\\
    BEVFormer \cite{li2022bevformer}    & C & R101             & $900\times1600$& 46.7 & 77.5 & 1.7\\
    Simple-BEV \cite{harley2022simple}& C & R101             & $448\times800$ & 47.4 &  -   & -\\
    Simple-BEV \cite{harley2022simple}&C+R& R101             & $448\times800$ & 55.7 &  -   & 7.3\\
    \gr \textbf{CRN}                    &C+R& R50              & $256\times704$ & \textbf{58.8} & \textbf{82.1} & 24.8 \\
    \hline
\end{tabular}}
\end{center}
\vspace{-5pt}
\caption{
\textbf{BEV Segmentation} on nuScenes \texttt{val} set.
}
\vspace{-10pt}
\label{table:Seg val set}
\end{table}

\noindent\textbf{3D Object Tracking and BEV Segmentation.\hspace{0.3cm}}
We further demonstrate the generalization performance of CRN on 3D object tracking and BEV segmentation tasks.
As shown in Table~\ref{table:3D Track test set}, our tracking result outperforms all published camera-only methods on nuScenes \texttt{test} set.
Also, ours not only significantly improves AMOTA but also reduces AMOTP and identity switches.

CRN consistently achieves state-of-the-art performance on BEV segmentation task as shown in Table~\ref{table:Seg val set}.
When compared to previous segmentation methods with a small image input and backbone~\cite{zhou2022cross, chen2022efficient}, ours performs significantly better while maintaining a real-time inference speed thanks to our semantically rich and spatially accurate BEV feature map from camera and radar.
CRN also achieves higher performance at a much faster FPS than a large image input and backbone~\cite{li2022bevformer}, demonstrating the effectiveness of fusion.

\subsection{Ablation Studies}
We conduct ablation studies on \texttt{val} set with a 3D detection task.
Unless otherwise specified, models use two frames of $256\times704$ image, R50 backbone, and are trained for 24 epochs without CBGS~\cite{zhu2019class}.
For thorough comparison, we additionally build three baseline detectors for camera -- BEVDepth~\cite{li2022bevdepth}, point -- CenterPoint~\cite{yin2021center}, and camera-point -- BEVFusion~\cite{liu2022bevfusion}.
Details of baselines and additional ablation studies are provided in Appendix.

\setlength{\tabcolsep}{0.6em}
\begin{table}[!t]
\begin{center}
\resizebox{0.88\columnwidth}{!}{
\begin{tabular}{l|c||ccc|c}
\hline
\multirow{2}{*}{Input} & \multirow{2}{*}{RVT} & \multicolumn{3}{c|}{\textit{All}} & \textit{Car} \\
& & NDS & mAP & mATE & mAP \\
\hline
Depth & \small{\xmark} & 43.9 & 33.2 & 0.716 & 50.4 \\ 
Radar & \small{\xmark} & 33.6 & 24.3 & 0.706 & 44.7 \\
\hline
Depth+Radar & \small{\cmark} & 52.1 & 44.8 & 0.521 & 70.5 \\
Depth+LiDAR & \small{\cmark} & 57.0 & 51.6 & 0.419 & 76.2 \\
\hline
\end{tabular}}
\end{center}
\vspace{-5pt}
\caption{
Ablation of view transformation methods. 
LiDAR and radar are used only for transformation and not used for feature aggregation.
}
\vspace{-6pt}
\label{table:ablation_RVT}
\end{table}

\setlength{\tabcolsep}{0.6em}
\begin{table}[!t]
\begin{center}
\resizebox{0.88\columnwidth}{!}{
\begin{tabular}{l|c||ccc|c}
\hline
& \multirow{2}{*}{Input} & \multicolumn{3}{c|}{\textit{All}} & \textit{Car} \\
& & NDS & mAP & mATE & mAP \\
\hline
CenterPoint & {L}   & 52.8 & 41.2 & 0.406 & 73.9 \\
BEVDepth    & {C}   & 43.9 & 33.2 & 0.716 & 50.4 \\ 
\hline
BEVFusion     & {C+R} & 51.9 & 42.4 & 0.536 & 68.4 \\ 
+ deeper conv & {C+R} & 51.9 & 42.8 & 0.532 & 69.0 \\ 
+ RVT         & {C+R} & 52.7 & 44.3 & 0.517 & 70.6 \\ 
\hline
MFA           & {C+R} & 53.4 & 44.5 & 0.507 & 70.3 \\ 
+ RVT         & {C+R} & 53.9 & 45.2 & 0.501 & 71.6 \\ 
\hline
\end{tabular}}
\end{center}
\vspace{-5pt}
\caption{
Ablation of feature aggregation methods.
Note that MFA with RVT is our full model.
}
\label{table:ablation_MFA}
\end{table}

\vspace{3pt}
\noindent\textbf{View Transformation.\hspace{0.2cm}}
In Table~\ref{table:ablation_RVT}, we study how the radar-assisted feature transformation affects performance.
View transformation solely relying on estimated depth suffers from inaccurate localization due to the inherent low accuracy of depth distribution.
If we naively replace depth distribution to radar (1 if radar point exists inside the voxel, 0 else), performance is severely degraded.
This is because image features in perspective view cannot be properly transformed due to the ambiguity and sparsity of radar.
With the proposed RVT, the model can benefit from both dense depth and sparse range measurement to significantly improve performance (+8.2\% NDS, +11.6\% mAP) over depth-only transformation.
Moreover, we find consistent performance improvement on LiDAR input, showing the effectiveness of RVT.

\vspace{3pt}
\noindent\textbf{Feature Aggregation.\hspace{0.2cm}}
Table~\ref{table:ablation_MFA} shows the comparison between different feature aggregation methods.
BEVFusion~\cite{liu2022bevfusion} fuses multi-modal feature maps in BEV using a single convolutional layer, which is not adaptive and has a small receptive field ($3\times3$).
Simply adding two additional convolutional layers for fusion, which provides a larger receptive field ($7\times7$) and bigger capacity, does not improve the performance much.
On the other hand, using only MFA already outperforms deeper BEVFusion with RVT, showing the effectiveness of the proposed multi-modal deformable cross attention.
We find that the performance gain of RVT is less significant on MFA than BEVFusion since MFA is already capable of handling spatial misalignment between multi-modal features. 

\setlength{\tabcolsep}{0.5em}
\begin{table}[!t]
\begin{center}
\resizebox{0.94\columnwidth}{!}{
\begin{tabular}{l|c||cccc|c}
\hline
& \multirow{2}{*}{Input} & \multicolumn{4}{c|}{\textit{Car} mAP} &\multirow{2}{*}{FPS} \\
& & \small{[0,100)} & \small{[0,30)} & \small{[30,60)} & \small{[60,100)} &  \\
\hline
CenterPoint & L & 54.2 & \textbf{84.3} & 35.8 & 4.8 & 6.3 \\
\hline
BEVDepth    & C & 34.1 & 65.4 & 13.7 & 0.2 & 13.0 \\
CenterPoint & R & 20.3 & 36.6 & 11.6 & 2.9 & \textbf{30.7} \\
\hline
CRN        & C+R & \textbf{56.9} & 82.6 & \textbf{42.6} & \textbf{7.0} & 11.5 \\
CRN-S      & C+R & 54.0 & 79.2 & 39.8 & 6.2 & 14.0 \\
\hline
\end{tabular}}
\end{center}
\vspace{-5pt}
\caption{
Analysis over various perception ranges.
Suffix -S denotes sparse aggregation, and we use $256\times 704$ and R50 for all camera streams.
}
\vspace{-6pt}
\label{table:ablation_range}
\end{table}

\setlength{\tabcolsep}{0.4em}
\begin{table}[!t]
\begin{center}
\resizebox{0.96\columnwidth}{!}{
\begin{tabular}{l|c|c||cccc}
\hline
& \multirow{2}{*}{Input} & \multirow{2}{*}{Drop} & \multicolumn{4}{c}{\# of view drops} \\
&  &  & 0 & 1 & 3 & 6 \\
\hline
BEVDepth    &  C  & C & 49.4 & 41.1 & 24.2 & 0 \\
CenterPoint &  R  & R & 30.6 & 25.3 & 14.9 & 0 \\
\hline
\multirow{2}{*}{BEVFusion} & \multirow{2}{*}{C+R} & C & \multirow{2}{*}{63.9} 
& 58.5 & 45.7 & \textbf{14.3} \\
& & R & & 59.9 & 50.9 & 34.4 \\
\hline
\multirow{2}{*}{CRN}       & \multirow{2}{*}{C+R} & C & \multirow{2}{*}{\shortstack{\textbf{68.8}\\\scriptsize{(+4.9)}}} 
& \textbf{62.4}\scriptsize{(+3.9)} & \textbf{48.9}\scriptsize{(+3.2)} & 12.8\scriptsize{(-1.5)} \\
& & R & & \textbf{64.3}\scriptsize{(+4.4)} & \textbf{57.0}\scriptsize{(+6.1)} & \textbf{43.8}\scriptsize{(+9.4)} \\
\hline
\end{tabular}}
\end{center}
\vspace{-5pt}
\caption{
Analysis of robustness using \textit{Car} class mAP.
Six view drops denote the single modality is entirely off.
}
\label{table:analysis_robust}
\end{table}

\subsection{Analysis}
\noindent\textbf{Scaling Up Perception Range.\hspace{0.2cm}}
In Table~\ref{table:ablation_range}, we extend the perception range from 51.2$m$ to 102.4$m$ and also increase the evaluation range twice correspondingly (see Appendix for details).
Although CenterPoint~\cite{yin2021center} uses 10 LiDAR sweeps, points become extremely sparse as the range increases, and thus performance is significantly degraded at far distances.
On the other hand, CRN outperforms LiDAR especially at farther than 30$m$ range with a much faster FPS, showing the effectiveness of camera and radar for long-range perception.
Moreover, CRN with sparse aggregation further improves the inference speed while preserving comparable performance.

\vspace{3pt}
\noindent\textbf{Robustness.\hspace{0.2cm}}
To systematically analyze the robustness of sensor failure cases, we randomly drop image and radar inputs in Table~\ref{table:analysis_robust}.
For fair comparisons, we use \textit{single} frame input and fix the seed to ensure the same views can be dropped over experiments.
We also train both fusion methods with data-level augmentation~\cite{chen2022autoalignv2}.
CRN not only outperforms BEVFusion when all modalities are available but maintains higher mAP on sensor failure cases.
Considering that ours uses radar points at multiple stages (RVT and MFA), each proposed module is trained to be robust to sparse and ambiguous radar points.
Especially when radar input is entirely unavailable, BEVFusion suffers from a performance drop over BEVDepth (-15.0\%), while CRN still keeps the competitive performance (-5.6\%).
This advantage comes from our attention module, which can adaptively choose modalities to use.

\vspace{3pt}
\noindent\textbf{Weather and Lighting.\hspace{0.2cm}}
We analyze the performance under different weather and lightning conditions in Table~\ref{table:analysis_weather}.
Note that R101 backbone with $512\times1408$ input is used for BEVDepth and ours for comparable comparisons with LiDAR methods.
Sensor noises of LiDAR in rainy conditions or poor illumination of camera at night make object detection challenging for LiDAR-only or camera-only methods.
Thanks to fusion with radar, ours shows consistent performance improvement of more than 10 mAP over the camera-only method, demonstrating the effectiveness and robustness of camera and radar sensors in all weather conditions.

\setlength{\tabcolsep}{0.25em}
\begin{table}[!t]
\begin{center}
\resizebox{0.98\columnwidth}{!}{
\begin{tabular}{l|c||cccc}
\hline
& Input & Sunny & Rainy & Day & Night \\
\hline
CenterPoint~\cite{yin2021center}  & \small{L}         & 62.9 & 59.2 & 62.8 & 35.4 \\
\hline
RCBEV~\cite{zhou2023bridging}     & \small{C+R} & 36.1 & 38.5 & 37.1 & 15.5 \\
BEVDepth \cite{li2022bevdepth}   & \small{C} & 39.0 & 39.0 & 39.3 & 16.8 \\
CRN     & \small{C+R} & \textbf{54.8}\scriptsize{(+15.8)} & \textbf{57.0}\scriptsize{(+18.0)} & \textbf{55.1}\scriptsize{(+15.8)} & \textbf{30.4}\scriptsize{(+13.6)} \\
\hline
\end{tabular}}
\end{center}
\vspace{-5pt}
\caption{
Analysis of different lighting and weather conditions using mAP metric.
CenterPoint~\cite{yin2021center} results are from BEVFusion~\cite{liu2022bevfusion}, and BEVDepth results are reproduced by us.
}
\label{table:analysis_weather}
\vspace{-5pt}
\end{table}

\begin{figure}[t]
\begin{center}
\includegraphics[width=0.96\columnwidth]{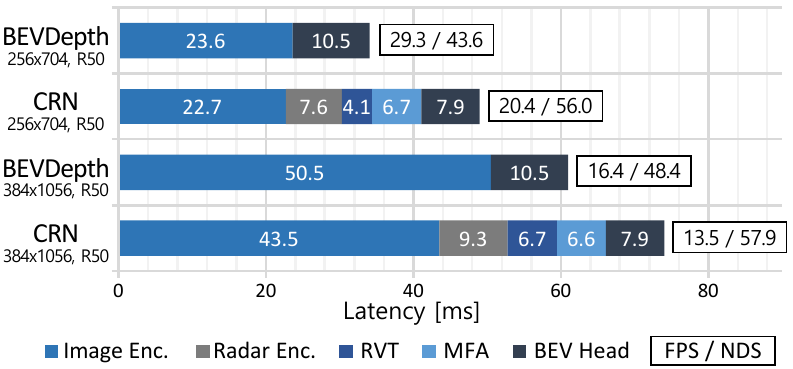}
\end{center}
\vspace{-8pt}
\caption{
    Inference time analysis of proposed components.
    All latency numbers are measured with batch size 1, GPU warm-up, and FP16 precision.
}
\vspace{-5pt}
\label{fig:analysis_speed}
\end{figure}

\vspace{3pt}
\noindent\textbf{Inference Time.\hspace{0.2cm}} \label{sec:inference time}
We analyze the inference time of each proposed component in Fig.~\ref{fig:analysis_speed}.
In all analyses, we assume that the BEV feature map of the previous frame $T-1$ can be stored and accessed at the current frame $T$ since ours does not use temporal information (\textit{e.g.}, temporal stereo methods~\cite{li2022bevstereo, park2022time}) when obtaining the BEV feature map.
It means that using a multi-frame only increases the latency of the BEV head.
Ours requires negligible additional computation for point encoder and fusion modules, but the performance gain over additional latency is substantial (+14.9$ms$ for +12.4 NDS in 256x704 and R50 setting).
Moreover, ours with small input can outperform camera-only with larger input in terms of both latency and performance.
We expect that inference optimization methods (\textit{e.g.}, TensorRT) can further reduce the latency of large model for long perception range setting to match the real-time.

\section{Conclusion}\label{sec:conclusion}

We present CRN, a novel camera-radar fusion method for accurate, robust, and efficient multi-task 3D perception. 
Our method effectively overcomes the limitation of each modality and fuses multi-modal information to generate contextually rich and spatially accurate BEV scene representation.
CRN is also suitable for long-range perception in real-time and achieves state-of-the-art performance on various tasks.
We hope that CRN will inspire future research on camera-radar fusion for 3D perception.

\section*{Acknowledgment}
This work was supported by Institute of Information \& communications Technology Planning \& Evaluation (IITP) and the National Research Foundation of Korea (NRF) grant funded by the Korea government (MSIT)
(RS-2023-00236245, Development of Perception/Planning AI SW for Seamless Autonomous Driving in Adverse Weather/Unstructured Environment and 2022R1A2C200494412)

{\small
\bibliographystyle{ieee_fullname}
\bibliography{egbib}

\begin{thebibliography}{10}\itemsep=-1pt

\bibitem{bai2022transfusion}
Xuyang Bai, Zeyu Hu, Xinge Zhu, Qingqiu Huang, Yilun Chen, Hongbo Fu, and
  Chiew-Lan Tai.
\newblock Transfusion: Robust lidar-camera fusion for 3d object detection with
  transformers.
\newblock In {\em Proceedings of the IEEE/CVF Conference on Computer Vision and
  Pattern Recognition (CVPR)}, pages 1090--1099, 2022.

\bibitem{Caesar2020}
Holger Caesar, Varun Bankiti, Alex~H. Lang, Sourabh Vora, Venice~Erin Liong,
  Qiang Xu, Anush Krishnan, Yu Pan, Giancarlo Baldan, and Oscar Beijbom.
\newblock nuscenes: A multimodal dataset for autonomous driving.
\newblock In {\em Proceedings of the IEEE/CVF Conference on Computer Vision and
  Pattern Recognition (CVPR)}, pages 11621--11631, 2020.

\bibitem{chaabane2021deft}
Mohamed Chaabane, Peter Zhang, J~Ross Beveridge, and Stephen O'Hara.
\newblock Deft: Detection embeddings for tracking.
\newblock In {\em Proceedings of the IEEE/CVF Conference on Computer Vision and
  Pattern Recognition Workshops (CVPRW)}, 2021.

\bibitem{chen2019mmdetection}
Kai Chen, Jiaqi Wang, Jiangmiao Pang, Yuhang Cao, Yu Xiong, Xiaoxiao Li,
  Shuyang Sun, Wansen Feng, Ziwei Liu, Jiarui Xu, et~al.
\newblock Mmdetection: Open mmlab detection toolbox and benchmark.
\newblock {\em arXiv preprint arXiv:1906.07155}, 2019.

\bibitem{chen2022efficient}
Shaoyu Chen, Tianheng Cheng, Xinggang Wang, Wenming Meng, Qian Zhang, and Wenyu
  Liu.
\newblock Efficient and robust 2d-to-bev representation learning via
  geometry-guided kernel transformer.
\newblock {\em arXiv preprint arXiv:2206.04584}, 2022.

\bibitem{chen2022autoalignv2}
Zehui Chen, Zhenyu Li, Shiquan Zhang, Liangji Fang, Qinhong Jiang, and Feng
  Zhao.
\newblock Autoalignv2: Deformable feature aggregation for dynamic multi-modal
  3d object detection.
\newblock In {\em Proceedings of the European Conference on Computer Vision
  (ECCV)}, pages 628--644, 2022.

\bibitem{cho2014multi}
Hyunggi Cho, Young-Woo Seo, BVK~Vijaya Kumar, and Ragunathan~Raj Rajkumar.
\newblock A multi-sensor fusion system for moving object detection and tracking
  in urban driving environments.
\newblock In {\em Proceedings of the IEEE International Conference on Robotics
  and Automation (ICRA)}, pages 1836--1843, 2014.

\bibitem{ars408}
Continental.
\newblock {Continental ARS 408-21 Datasheet}.
\newblock \url{https://conti-engineering.com/components/ars-408/}.
\newblock Accessed: 2023-03-01.

\bibitem{mmcv}
MMCV Contributors.
\newblock {MMCV: OpenMMLab} computer vision foundation.
\newblock \url{https://github.com/open-mmlab/mmcv}, 2018.

\bibitem{deng2009imagenet}
Jia Deng, Wei Dong, Richard Socher, Li-Jia Li, Kai Li, and Li Fei-Fei.
\newblock Imagenet: A large-scale hierarchical image database.
\newblock In {\em Proceedings of the IEEE/CVF Conference on Computer Vision and
  Pattern Recognition (CVPR)}, pages 248--255, 2009.

\bibitem{dong2022superfusion}
Hao Dong, Xianjing Zhang, Xuan Jiang, Jun Zhang, Jintao Xu, Rui Ai, Weihao Gu,
  Huimin Lu, Juho Kannala, and Xieyuanli Chen.
\newblock Superfusion: Multilevel lidar-camera fusion for long-range hd map
  generation and prediction.
\newblock {\em arXiv preprint arXiv:2211.15656}, 2022.

\bibitem{drews2022deepfusion}
Florian Drews, Di Feng, Florian Faion, Lars Rosenbaum, Michael Ulrich, and
  Claudius Gl{\"a}ser.
\newblock Deepfusion: A robust and modular 3d object detector for lidars,
  cameras and radars.
\newblock In {\em Proceedings of the IEEE/RSJ International Conference on
  Intelligent Robots and Systems (IROS)}, pages 560--567, 2022.

\bibitem{everingham2010pascal}
Mark Everingham, Luc Van~Gool, Christopher~KI Williams, John Winn, and Andrew
  Zisserman.
\newblock The pascal visual object classes (voc) challenge.
\newblock {\em International Journal of Computer Vision (Int. J. Comput.
  Vis.)}, 88(2):303--338, 2010.

\bibitem{fischer2022cc}
Tobias Fischer, Yung-Hsu Yang, Suryansh Kumar, Min Sun, and Fisher Yu.
\newblock Cc-3dt: Panoramic 3d object tracking via cross-camera fusion.
\newblock In {\em Proceedings of the Conference on Robot Learning (CoRL)},
  pages 2294--2305, 2023.

\bibitem{gohring2011radar}
Daniel G{\"o}hring, Miao Wang, Michael Schn{\"u}rmacher, and Tinosch Ganjineh.
\newblock Radar/lidar sensor fusion for car-following on highways.
\newblock In {\em Proceedings of the IEEE International Conference on
  Automation, Robotics and Applications (ICARA)}, pages 407--412, 2011.

\bibitem{Guizilini2020}
Vitor Guizilini, Rares Ambrus, Sudeep Pillai, Allan Raventos, and Adrien
  Gaidon.
\newblock {3D Packing for Self-Supervised Monocular Depth Estimation}.
\newblock In {\em Proceedings of the IEEE/CVF Conference on Computer Vision and
  Pattern Recognition (CVPR)}, pages 2485--2494, 2020.

\bibitem{harley2022simple}
Adam~W Harley, Zhaoyuan Fang, Jie Li, Rares Ambrus, and Katerina Fragkiadaki.
\newblock Simple-bev: What really matters for multi-sensor bev perception?
\newblock In {\em Proceedings of the IEEE International Conference on Robotics
  and Automation (ICRA)}, pages 2759--2765, 2023.

\bibitem{he2016deep}
Kaiming He, Xiangyu Zhang, Shaoqing Ren, and Jian Sun.
\newblock Deep residual learning for image recognition.
\newblock In {\em Proceedings of the IEEE/CVF Conference on Computer Vision and
  Pattern Recognition (CVPR)}, pages 770--778, 2016.

\bibitem{hu2021fiery}
Anthony Hu, Zak Murez, Nikhil Mohan, Sof{\'\i}a Dudas, Jeffrey Hawke, Vijay
  Badrinarayanan, Roberto Cipolla, and Alex Kendall.
\newblock Fiery: Future instance prediction in bird's-eye view from surround
  monocular cameras.
\newblock In {\em Proceedings of the IEEE/CVF International Conference on
  Computer Vision (ICCV)}, pages 15273--15282, 2021.

\bibitem{hu2022monocular}
Hou-Ning Hu, Yung-Hsu Yang, Tobias Fischer, Trevor Darrell, Fisher Yu, and Min
  Sun.
\newblock Monocular quasi-dense 3d object tracking.
\newblock {\em IEEE Transactions on Pattern Analysis and Machine Intelligence
  (IEEE Trans. Pattern Anal. Mach. Intell.)}, 45(2):1992--2008, 2022.

\bibitem{huang2016deep}
Gao Huang, Yu Sun, Zhuang Liu, Daniel Sedra, and Kilian~Q Weinberger.
\newblock Deep networks with stochastic depth.
\newblock In {\em Proceedings of the European Conference on Computer Vision
  (ECCV)}, pages 646--661, 2016.

\bibitem{huang2021bevdet}
Junjie Huang, Guan Huang, Zheng Zhu, and Dalong Du.
\newblock Bevdet: High-performance multi-camera 3d object detection in
  bird-eye-view.
\newblock In {\em arXiv preprint arXiv:2112.11790}, 2021.

\bibitem{jiang2022polarformer}
Yanqin Jiang, Li Zhang, Zhenwei Miao, Xiatian Zhu, Jin Gao, Weiming Hu, and
  Yu-Gang Jiang.
\newblock Polarformer: Multi-camera 3d object detection with polar
  transformers.
\newblock In {\em Proceedings of the AAAI Conference on Artificial Intelligence
  (AAAI)}, 2023.

\bibitem{johnson1992array}
Don~H Johnson and Dan~E Dudgeon.
\newblock {\em Array signal processing: concepts and techniques}.
\newblock Simon \& Schuster, Inc., 1992.

\bibitem{kim2020low}
Jinhyeong Kim, Youngseok Kim, and Dongsuk Kum.
\newblock Low-level sensor fusion network for 3d vehicle detection using radar
  range-azimuth heatmap and monocular image.
\newblock In {\em Proceedings of the Asian Conference on Computer Vision
  (ACCV)}, pages 388--402, 2020.

\bibitem{kim2020grif}
Youngseok Kim, Jun~Won Choi, and Dongsuk Kum.
\newblock {GRIF Net: Gated region of interest fusion network for robust 3D
  object detection from radar point cloud and monocular image}.
\newblock In {\em Proceedings of the IEEE/RSJ International Conference on
  Intelligent Robots and Systems (IROS)}, pages 10857--10864, 2020.

\bibitem{kim2022craft}
Youngseok Kim, Sanmin Kim, Jun~Won Choi, and Dongsuk Kum.
\newblock {CRAFT: Camera-Radar 3D Object Detection with Spatio-Contextual
  Fusion Transformer}.
\newblock In {\em Proceedings of the AAAI Conference on Artificial Intelligence
  (AAAI)}, 2023.

\bibitem{ku2018joint}
Jason Ku, Melissa Mozifian, Jungwook Lee, Ali Harakeh, and Steven~L Waslander.
\newblock Joint 3d proposal generation and object detection from view
  aggregation.
\newblock In {\em Proceedings of the IEEE/RSJ International Conference on
  Intelligent Robots and Systems (IROS)}, pages 5750--5757, 2018.

\bibitem{lang2019pointpillars}
Alex~H Lang, Sourabh Vora, Holger Caesar, Lubing Zhou, Jiong Yang, and Oscar
  Beijbom.
\newblock Pointpillars: Fast encoders for object detection from point clouds.
\newblock In {\em Proceedings of the IEEE/CVF Conference on Computer Vision and
  Pattern Recognition (CVPR)}, pages 12697--12705, 2019.

\bibitem{leng2022lidaraugment}
Zhaoqi Leng, Guowang Li, Chenxi Liu, Ekin~Dogus Cubuk, Pei Sun, Tong He,
  Dragomir Anguelov, and Mingxing Tan.
\newblock Lidaraugment: Searching for scalable 3d lidar data augmentations.
\newblock In {\em Proceedings of the IEEE International Conference on Robotics
  and Automation (ICRA)}, pages 7039--7045, 2023.

\bibitem{li2008mimo}
Jian Li and Petre Stoica.
\newblock {\em MIMO radar signal processing}.
\newblock John Wiley \& Sons, 2008.

\bibitem{li2022hdmapnet}
Qi Li, Yue Wang, Yilun Wang, and Hang Zhao.
\newblock Hdmapnet: An online hd map construction and evaluation framework.
\newblock In {\em Proceedings of the IEEE International Conference on Robotics
  and Automation (ICRA)}, pages 4628--4634, 2022.

\bibitem{li2022bevstereo}
Yinhao Li, Han Bao, Zheng Ge, Jinrong Yang, Jianjian Sun, and Zeming Li.
\newblock Bevstereo: Enhancing depth estimation in multi-view 3d object
  detection with dynamic temporal stereo.
\newblock In {\em Proceedings of the AAAI Conference on Artificial Intelligence
  (AAAI)}, 2023.

\bibitem{li2022unifying}
Yanwei Li, Yilun Chen, Xiaojuan Qi, Zeming Li, Jian Sun, and Jiaya Jia.
\newblock Unifying voxel-based representation with transformer for 3d object
  detection.
\newblock In {\em Advances in Neural Information Processing Systems (NeurIPS)},
  2022.

\bibitem{li2022bevdepth}
Yinhao Li, Zheng Ge, Guanyi Yu, Jinrong Yang, Zengran Wang, Yukang Shi,
  Jianjian Sun, and Zeming Li.
\newblock Bevdepth: Acquisition of reliable depth for multi-view 3d object
  detection.
\newblock In {\em Proceedings of the AAAI Conference on Artificial Intelligence
  (AAAI)}, 2023.

\bibitem{li2022bevformer}
Zhiqi Li, Wenhai Wang, Hongyang Li, Enze Xie, Chonghao Sima, Tong Lu, Qiao Yu,
  and Jifeng Dai.
\newblock Bevformer: Learning bird's-eye-view representation from multi-camera
  images via spatiotemporal transformers.
\newblock In {\em Proceedings of the European Conference on Computer Vision
  (ECCV)}, pages 1--18, 2022.

\bibitem{liang2022bevfusion}
Tingting Liang, Hongwei Xie, Kaicheng Yu, Zhongyu Xia, Zhiwei Lin, Yongtao
  Wang, Tao Tang, Bing Wang, and Zhi Tang.
\newblock Bevfusion: A simple and robust lidar-camera fusion framework.
\newblock In {\em Advances in Neural Information Processing Systems (NeurIPS)},
  2022.

\bibitem{lin2020depth}
Juan-Ting Lin, Dengxin Dai, and Luc Van~Gool.
\newblock Depth estimation from monocular images and sparse radar data.
\newblock In {\em Proceedings of the IEEE/RSJ International Conference on
  Intelligent Robots and Systems (IROS)}, pages 10233--10240, 2020.

\bibitem{lin2017feature}
Tsung-Yi Lin, Piotr Doll{\'a}r, Ross Girshick, Kaiming He, Bharath Hariharan,
  and Serge Belongie.
\newblock Feature pyramid networks for object detection.
\newblock In {\em Proceedings of the IEEE/CVF Conference on Computer Vision and
  Pattern Recognition (CVPR)}, pages 2117--2125, 2017.

\bibitem{Lin2017a}
Tsung-yi Lin, Priya Goyal, Ross Girshick, Kaiming He, and Piotr Dollar.
\newblock {Focal Loss for Dense Object Detection}.
\newblock In {\em Proceedings of the IEEE/CVF International Conference on
  Computer Vision (ICCV)}, pages 2980--2988, 2017.

\bibitem{lin2022sparse4d}
Xuewu Lin, Tianwei Lin, Zixiang Pei, Lichao Huang, and Zhizhong Su.
\newblock Sparse4d: Multi-view 3d object detection with sparse spatial-temporal
  fusion.
\newblock {\em arXiv preprint arXiv:2211.10581}, 2022.

\bibitem{lin2018human}
Yier Lin, Julien Le~Kernec, Shufan Yang, Francesco Fioranelli, Olivier Romain,
  and Zhiqin Zhao.
\newblock Human activity classification with radar: Optimization and noise
  robustness with iterative convolutional neural networks followed with random
  forests.
\newblock {\em IEEE Sensors Journal}, 18(23):9669--9681, 2018.

\bibitem{liu2022petr}
Yingfei Liu, Tiancai Wang, Xiangyu Zhang, and Jian Sun.
\newblock Petr: Position embedding transformation for multi-view 3d object
  detection.
\newblock In {\em Proceedings of the European Conference on Computer Vision
  (ECCV)}, pages 531--–548, 2022.

\bibitem{liu2022convnet}
Zhuang Liu, Hanzi Mao, Chao-Yuan Wu, Christoph Feichtenhofer, Trevor Darrell,
  and Saining Xie.
\newblock A convnet for the 2020s.
\newblock In {\em Proceedings of the IEEE/CVF Conference on Computer Vision and
  Pattern Recognition (CVPR)}, pages 11976--11986, 2022.

\bibitem{liu2022bevfusion}
Zhijian Liu, Haotian Tang, Alexander Amini, Xinyu Yang, Huizi Mao, Daniela Rus,
  and Song Han.
\newblock Bevfusion: Multi-task multi-sensor fusion with unified bird's-eye
  view representation.
\newblock In {\em Proceedings of the IEEE International Conference on Robotics
  and Automation (ICRA)}, 2023.

\bibitem{long2021radar}
Yunfei Long, Daniel Morris, Xiaoming Liu, Marcos Castro, Punarjay Chakravarty,
  and Praveen Narayanan.
\newblock Radar-camera pixel depth association for depth completion.
\newblock In {\em Proceedings of the IEEE/CVF Conference on Computer Vision and
  Pattern Recognition (CVPR)}, pages 12507--12516, 2021.

\bibitem{loshchilov2018decoupled}
Ilya Loshchilov and Frank Hutter.
\newblock Decoupled weight decay regularization.
\newblock In {\em Proceedings of the International Conference on Learning
  Representations (ICLR)}, 2019.

\bibitem{lu2022learning}
Jiachen Lu, Zheyuan Zhou, Xiatian Zhu, Hang Xu, and Li Zhang.
\newblock Learning ego 3d representation as ray tracing.
\newblock In {\em Proceedings of the European Conference on Computer Vision
  (ECCV)}, pages 129--144, 2022.

\bibitem{major2019vehicle}
Bence Major, Daniel Fontijne, Amin Ansari, Ravi Teja~Sukhavasi, Radhika
  Gowaikar, Michael Hamilton, Sean Lee, Slawomir Grzechnik, and Sundar
  Subramanian.
\newblock Vehicle detection with automotive radar using deep learning on
  range-azimuth-doppler tensors.
\newblock In {\em Proceedings of the IEEE/CVF International Conference on
  Computer Vision Workshops (ICCVW)}, pages 924--932, 2019.

\bibitem{meyer2019automotive}
Michael Meyer and Georg Kuschk.
\newblock Automotive radar dataset for deep learning based 3d object detection.
\newblock In {\em Proceedings of the European Radar Conference (EuRAD)}, pages
  129--132, 2019.

\bibitem{nabati2021centerfusion}
Ramin Nabati and Hairong Qi.
\newblock Centerfusion: Center-based radar and camera fusion for 3d object
  detection.
\newblock In {\em Proceedings of the IEEE/CVF Winter Conference on Applications
  of Computer Vision (WACV)}, pages 1527--1536, 2021.

\bibitem{pang2020clocs}
Su Pang, Daniel Morris, and Hayder Radha.
\newblock Clocs: Camera-lidar object candidates fusion for 3d object detection.
\newblock In {\em Proceedings of the IEEE/RSJ International Conference on
  Intelligent Robots and Systems (IROS)}, pages 10386--10393. IEEE, 2020.

\bibitem{park2021pseudo}
Dennis Park, Rares Ambrus, Vitor Guizilini, Jie Li, and Adrien Gaidon.
\newblock Is pseudo-lidar needed for monocular 3d object detection?
\newblock In {\em Proceedings of the IEEE/CVF International Conference on
  Computer Vision (ICCV)}, pages 3142--3152, 2021.

\bibitem{park2022time}
Jinhyung Park, Chenfeng Xu, Shijia Yang, Kurt Keutzer, Kris Kitani, Masayoshi
  Tomizuka, and Wei Zhan.
\newblock Time will tell: New outlooks and a baseline for temporal multi-view
  3d object detection.
\newblock In {\em Proceedings of the International Conference on Learning
  Representations (ICLR)}, 2023.

\bibitem{philion2020lift}
Jonah Philion and Sanja Fidler.
\newblock Lift, splat, shoot: Encoding images from arbitrary camera rigs by
  implicitly unprojecting to 3d.
\newblock In {\em Proceedings of the European Conference on Computer Vision
  (ECCV)}, pages 194--210, 2020.

\bibitem{popov2022nvradarnet}
Alexander Popov, Patrik Gebhardt, Ke Chen, Ryan Oldja, Heeseok Lee, Shane
  Murray, Ruchi Bhargava, and Nikolai Smolyanskiy.
\newblock Nvradarnet: Real-time radar obstacle and free space detection for
  autonomous driving.
\newblock In {\em Proceedings of the IEEE International Conference on Robotics
  and Automation (ICRA)}, pages 6958--6964, 2023.

\bibitem{qi2018frustum}
Charles~R Qi, Wei Liu, Chenxia Wu, Hao Su, and Leonidas~J Guibas.
\newblock Frustum pointnets for 3d object detection from rgb-d data.
\newblock In {\em Proceedings of the IEEE/CVF Conference on Computer Vision and
  Pattern Recognition (CVPR)}, pages 918--927, 2018.

\bibitem{qi2017pointnet}
Charles~R Qi, Hao Su, Kaichun Mo, and Leonidas~J Guibas.
\newblock Pointnet: Deep learning on point sets for 3d classification and
  segmentation.
\newblock In {\em Proceedings of the IEEE/CVF Conference on Computer Vision and
  Pattern Recognition (CVPR)}, pages 652--660, 2017.

\bibitem{qi2017pointnet++}
Charles~Ruizhongtai Qi, Li Yi, Hao Su, and Leonidas~J Guibas.
\newblock Pointnet++: Deep hierarchical feature learning on point sets in a
  metric space.
\newblock In {\em Advances in Neural Information Processing Systems (NeurIPS)},
  pages 5105--5114, 2017.

\bibitem{reading2021categorical}
Cody Reading, Ali Harakeh, Julia Chae, and Steven~L Waslander.
\newblock Categorical depth distribution network for monocular 3d object
  detection.
\newblock In {\em Proceedings of the IEEE/CVF Conference on Computer Vision and
  Pattern Recognition (CVPR)}, pages 8555--8564, 2021.

\bibitem{roddick2020predicting}
Thomas Roddick and Roberto Cipolla.
\newblock Predicting semantic map representations from images using pyramid
  occupancy networks.
\newblock In {\em Proceedings of the IEEE/CVF Conference on Computer Vision and
  Pattern Recognition (CVPR)}, pages 11138--11147, 2020.

\bibitem{roh2021sparse}
Byungseok Roh, JaeWoong Shin, Wuhyun Shin, and Saehoon Kim.
\newblock Sparse detr: Efficient end-to-end object detection with learnable
  sparsity.
\newblock In {\em Proceedings of the International Conference on Learning
  Representations (ICLR)}, 2022.

\bibitem{saha2022translating}
Avishkar Saha, Oscar Mendez, Chris Russell, and Richard Bowden.
\newblock Translating images into maps.
\newblock In {\em Proceedings of the IEEE International Conference on Robotics
  and Automation (ICRA)}, pages 9200--9206, 2022.

\bibitem{Shi2020a}
Shaoshuai Shi, Chaoxu Guo, Li Jiang, Zhe Wang, Jianping Shi, Xiaogang Wang, and
  Hongsheng Li.
\newblock {PV-RCNN: Point-Voxel Feature Set Abstraction for 3D Object
  Detection}.
\newblock In {\em Proceedings of the IEEE/CVF Conference on Computer Vision and
  Pattern Recognition (CVPR)}, pages 10529--10538, 2020.

\bibitem{shi2019pointrcnn}
Shaoshuai Shi, Xiaogang Wang, and Hongsheng Li.
\newblock Pointrcnn: 3d object proposal generation and detection from point
  cloud.
\newblock In {\em Proceedings of the IEEE/CVF Conference on Computer Vision and
  Pattern Recognition (CVPR)}, pages 770--779, 2019.

\bibitem{shi2020point}
Weijing Shi and Raj Rajkumar.
\newblock Point-gnn: Graph neural network for 3d object detection in a point
  cloud.
\newblock In {\em Proceedings of the IEEE/CVF Conference on Computer Vision and
  Pattern Recognition (CVPR)}, pages 1711--1719, 2020.

\bibitem{shin2023instagram}
Juyeb Shin, Francois Rameau, Hyeonjun Jeong, and Dongsuk Kum.
\newblock {InstaGraM: Instance-level Graph Modeling for Vectorized HD Map
  Learning}.
\newblock {\em arXiv preprint arXiv:2301.04470}, 2023.

\bibitem{Simonelli}
Andrea Simonelli, Samuel Rota~Rota Bul{\`{o}}, Lorenzo Porzi, Manuel
  L{\'{o}}pez-Antequera, and Peter Kontschieder.
\newblock {Disentangling Monocular 3D Object Detection}.
\newblock In {\em Proceedings of the IEEE/CVF International Conference on
  Computer Vision (ICCV)}, pages 1991--1999, 2019.

\bibitem{sless2019road}
Liat Sless, Bat El~Shlomo, Gilad Cohen, and Shaul Oron.
\newblock Road scene understanding by occupancy grid learning from sparse radar
  clusters using semantic segmentation.
\newblock In {\em Proceedings of the IEEE/CVF International Conference on
  Computer Vision Workshops (ICCVW)}, 2019.

\bibitem{sun2021rsn}
Pei Sun, Weiyue Wang, Yuning Chai, Gamaleldin Elsayed, Alex Bewley, Xiao Zhang,
  Cristian Sminchisescu, and Dragomir Anguelov.
\newblock Rsn: Range sparse net for efficient, accurate lidar 3d object
  detection.
\newblock In {\em Proceedings of the IEEE/CVF Conference on Computer Vision and
  Pattern Recognition (CVPR)}, pages 5725--5734, 2021.

\bibitem{svenningsson2021radar}
Peter Svenningsson, Francesco Fioranelli, and Alexander Yarovoy.
\newblock Radar-pointgnn: Graph based object recognition for unstructured radar
  point-cloud data.
\newblock In {\em Proceedings of the IEEE Radar Conference (RadarConf)}, pages
  1--6, 2021.

\bibitem{thomas2019kpconv}
Hugues Thomas, Charles~R Qi, Jean-Emmanuel Deschaud, Beatriz Marcotegui,
  Fran{\c{c}}ois Goulette, and Leonidas~J Guibas.
\newblock Kpconv: Flexible and deformable convolution for point clouds.
\newblock In {\em Proceedings of the IEEE/CVF International Conference on
  Computer Vision (ICCV)}, pages 6411--6420, 2019.

\bibitem{tian2019fcos}
Zhi Tian, Chunhua Shen, Hao Chen, and Tong He.
\newblock Fcos: Fully convolutional one-stage object detection.
\newblock In {\em Proceedings of the IEEE/CVF International Conference on
  Computer Vision (ICCV)}, pages 9627--9636, 2019.

\bibitem{touvron2021going}
Hugo Touvron, Matthieu Cord, Alexandre Sablayrolles, Gabriel Synnaeve, and
  Herv{\'e} J{\'e}gou.
\newblock Going deeper with image transformers.
\newblock In {\em Proceedings of the IEEE/CVF International Conference on
  Computer Vision (ICCV)}, pages 32--42, 2021.

\bibitem{ulrich2022improved}
Michael Ulrich, Sascha Braun, Daniel K{\"o}hler, Daniel Niederl{\"o}hner,
  Florian Faion, Claudius Gl{\"a}ser, and Holger Blume.
\newblock Improved orientation estimation and detection with hybrid object
  detection networks for automotive radar.
\newblock In {\em Proceedings of the IEEE International Intelligent
  Transportation Systems Conference (ITSC)}, pages 111--117, 2022.

\bibitem{vaswani2017attention}
Ashish Vaswani, Noam Shazeer, Niki Parmar, Jakob Uszkoreit, Llion Jones,
  Aidan~N Gomez, {\L}ukasz Kaiser, and Illia Polosukhin.
\newblock Attention is all you need.
\newblock In {\em Advances in Neural Information Processing Systems (NeurIPS)},
  pages 6000--6010, 2017.

\bibitem{vora2020pointpainting}
Sourabh Vora, Alex~H Lang, Bassam Helou, and Oscar Beijbom.
\newblock Pointpainting: Sequential fusion for 3d object detection.
\newblock In {\em Proceedings of the IEEE/CVF Conference on Computer Vision and
  Pattern Recognition (CVPR)}, pages 4604--4612, 2020.

\bibitem{wang2021fcos3d}
Tai Wang, Xinge Zhu, Jiangmiao Pang, and Dahua Lin.
\newblock Fcos3d: Fully convolutional one-stage monocular 3d object detection.
\newblock In {\em Proceedings of the IEEE/CVF International Conference on
  Computer Vision Workshops (ICCVW)}, pages 913--922, 2021.

\bibitem{wang2020pillar}
Yue Wang, Alireza Fathi, Abhijit Kundu, David~A Ross, Caroline Pantofaru, Tom
  Funkhouser, and Justin Solomon.
\newblock Pillar-based object detection for autonomous driving.
\newblock In {\em Proceedings of the European Conference on Computer Vision
  (ECCV)}, pages 18--34, 2020.

\bibitem{weng2019baseline}
Xinshuo Weng and Kris Kitani.
\newblock A baseline for 3d multi-object tracking.
\newblock In {\em arXiv preprint arXiv:1907.03961}, 2019.

\bibitem{weston2019probably}
Rob Weston, Sarah Cen, Paul Newman, and Ingmar Posner.
\newblock Probably unknown: Deep inverse sensor modelling radar.
\newblock In {\em Proceedings of the IEEE International Conference on Robotics
  and Automation (ICRA)}, pages 5446--5452, 2019.

\bibitem{wu2023mvfusion}
Zizhang Wu, Guilian Chen, Yuanzhu Gan, Lei Wang, and Jian Pu.
\newblock Mvfusion: Multi-view 3d object detection with semantic-aligned radar
  and camera fusion.
\newblock In {\em Proceedings of the IEEE International Conference on Robotics
  and Automation (ICRA)}, 2023.

\bibitem{yan2018second}
Yan Yan, Yuxing Mao, and Bo Li.
\newblock Second: Sparsely embedded convolutional detection.
\newblock {\em Sensors}, 18(10):3337--3352, 2018.

\bibitem{yin2021center}
Tianwei Yin, Xingyi Zhou, and Philipp Krahenbuhl.
\newblock Center-based 3d object detection and tracking.
\newblock In {\em Proceedings of the IEEE/CVF Conference on Computer Vision and
  Pattern Recognition (CVPR)}, pages 11784--11793, 2021.

\bibitem{yoo20203d}
Jin~Hyeok Yoo, Yecheol Kim, Jisong Kim, and Jun~Won Choi.
\newblock 3d-cvf: Generating joint camera and lidar features using cross-view
  spatial feature fusion for 3d object detection.
\newblock In {\em Proceedings of the European Conference on Computer Vision
  (ECCV)}, pages 720--736, 2020.

\bibitem{zhang2020exploring}
Wenwei Zhang, Zhe Wang, and Chen~Change Loy.
\newblock Exploring data augmentation for multi-modality 3d object detection.
\newblock {\em arXiv preprint arXiv:2012.12741}, 2020.

\bibitem{zhou2022cross}
Brady Zhou and Philipp Kr{\"a}henb{\"u}hl.
\newblock Cross-view transformers for real-time map-view semantic segmentation.
\newblock In {\em Proceedings of the IEEE/CVF Conference on Computer Vision and
  Pattern Recognition (CVPR)}, pages 13760--13769, 2022.

\bibitem{zhou2023bridging}
Taohua Zhou, Junjie Chen, Yining Shi, Kun Jiang, Mengmeng Yang, and Diange
  Yang.
\newblock Bridging the view disparity between radar and camera features for
  multi-modal fusion 3d object detection.
\newblock {\em IEEE Transactions on Intelligent Vehicles (IEEE Trans. Intell.
  Veh.)}, 2023.

\bibitem{Zhou2019}
Xingyi Zhou, Dequan Wang, and Philipp Kr{\"a}henb{\"u}hl.
\newblock Objects as points.
\newblock In {\em arXiv preprint arXiv:1904.07850}, 2019.

\bibitem{zhu2019class}
Benjin Zhu, Zhengkai Jiang, Xiangxin Zhou, Zeming Li, and Gang Yu.
\newblock Class-balanced grouping and sampling for point cloud 3d object
  detection.
\newblock In {\em arXiv preprint arXiv:1908.09492}, 2019.

\bibitem{zhu2020deformable}
Xizhou Zhu, Weijie Su, Lewei Lu, Bin Li, Xiaogang Wang, and Jifeng Dai.
\newblock Deformable detr: Deformable transformers for end-to-end object
  detection.
\newblock In {\em Proceedings of the International Conference on Learning
  Representations (ICLR)}, 2021.

\end{thebibliography}
}

\clearpage
{\Huge \textbf{Appendix}}
\setcounter{figure}{6}
\setcounter{table}{9}


\renewcommand\thesection{\Alph{section}}
\setcounter{section}{0}
\section{Overview}
This supplementary material provides additional details of architecture, qualitative and quantitative experimental results.
We describe the notation of MDCA (Sec.~\ref{appendix:notation}) and implementation details for experiments in the main paper (Sec.~\ref{appendix:implementation}).
We further provide additional experimental results (Sec.~\ref{appendix:experiment}) and qualitative results (Sec.~\ref{appendix:qualitative}).

\section{Multi-modal Deformable Cross Attention} \label{appendix:notation}
We adopt the deformable attention~\cite{zhu2020deformable} and extend it for multi-modal feature maps, denoted as multi-modal deformable cross attention (MDCA).

Given an input queries $\mathrm{z}_q$ and flattened multi-modal BEV feature maps $\mathrm{x}_m=\{\mathbf{C}_{I}^{BEV}, \mathbf{C}_{R}^{BEV} \in \mathbb{R}^{C\times XY}\}$, let $q$ index a query element and $p_q \in [0,1]^2$ be the normalized coordinates of the reference point for each query element $q$.
The multi-modal deformable cross attention (MDCA) is defined as
\setcounter{equation}{7}
\begin{equation}
\begin{split}
    & \text{MDCA}(\mathrm{z}_q,p_q,\mathrm{x}_m)= \\
    & \sum_{h}^{H} \boldsymbol{W}_h \left[ \sum_{m}^{M} \sum_{k}^{K} A_{hmqk} \cdot \boldsymbol{W}'_{hm} \mathrm{x}_m (\phi_{m}(p_q + \Delta p_{hmqk})) \right].
\end{split}
\end{equation}
$h, m, k$ index the attention head $H$, multiple modalities $\{\mathbf{C}_{I}, \mathbf{C}_{R}\}$, and the number of sampling points $K$.
$\boldsymbol{W}_h \in \mathbb{R}^{C\times C_v}$ is the output projection matrix at $h^{th}$ head, and $\boldsymbol{W}'_{hm} \in \mathbb{R}^{C_v\times C}$ is the input value projection matrix at $h^{th}$ head and modality $m$.
We use $C_v = C/H$ following multi-head attention in Transformers~\cite{vaswani2017attention}.
Note that separated input value projection matrices $\boldsymbol{W}'_{hm}$ are used for each modality to make MDCA modality-specific and achieve robust fusion (\textit{e.g.}, sensor failure case).
Both $A_{hmqk}$ and $\Delta p_{hmqk}$ are obtained by linear projection over the input queries $\mathrm{z}_q$, and the attention weight $A_{hmqk}$ is normalized to modalities and sampling points as $\sum\nolimits_m^M \sum\nolimits_k^K A_{hmqk} = 1$.
Function $\phi_{m}(p_q)$ scales the normalized coordinates $p_q$ in case two modalities have different shapes.

The proposed multi-modal deformable attention module is designed to look over multi-modal feature maps and multiple sampling points.
This can overcome spatial misalignment around reference points and enable adaptive fusion over modalities.

\section{Implementation Details} \label{appendix:implementation}
This section provides the experimental settings for the main results and ablation studies.

\subsection{Pre-processing and Hyper-parameters}
For the camera stream, the image backbone yields 4 levels of feature maps of stride 4, 8, 16, and 32, and we employ SECONDFPN~\cite{yan2018second}, which concatenates output feature maps at stride 16.
\texttt{nn.Conv2d} and \texttt{nn.ConvTranspose2d} are used for downsampling and upsampling.
Given FPN feature maps, the depth distribution network outputs $D$ size depth bins.
We use uniform discretization with a depth range of $[2.0, 58.0] m$ and bin size of $0.5 m$, resulting in $D=112$.

As stated in the main paper, we first project point cloud into an image coordinate system while preserving its depth and features for radar stream.
Note that the projection matrix for radar point projection corresponds to the image stream.
Next, we voxelize radar points in the frustum coordinate system $(d,u,v)$ to have the same size with an image frustum feature.
Taking into account the sparsity and accuracy of radar, we use $8\times$ downsampled pillar canvas and further extract pillar features using SECOND backbone, which yields 3 levels of feature maps of stride of 1, 2, and 4.
Finally, SECONDFPN is employed to pillar feature maps to output $16\times$ downsampled size in the image width direction and to have $D=112$ in a depth direction.

We use \texttt{multi\_scale\_deform\_attn} implementation from MMCV~\cite{mmcv} for deformable cross attention in Multi-modal Feature Aggregation (MFA).
Specifically, we use 6 layers of MFA, 8 attention heads, and 4 sampling points for MFA in our experiments.
MFA is applied to the single-frame camera and radar inputs and produces a fused BEV feature map.
After, fused BEV feature maps from the previous $T$ timestamps are aligned to the current timestamp and concatenated.
We use $T=3$ for the submission and $T=1$ for ablation studies and note that future frames are not used.

Following standard practices in monocular 3D object detection~\cite{huang2021bevdet,li2022bevdepth}, we set perception range $[-51.2, 51.2] m$ with a pillar size of $(0.2, 0.2) m$ and a downsampling factor of 4.
As a result, the size of the BEV feature map is $128\times128$.

For the BEV segmentation task, the perception range is set to $[-50.0, 50.0] m$ in both $X-$ and $Y-$axis centered around the ego vehicle, following previous works \cite{philion2020lift,hu2021fiery,zhou2022cross,chen2022efficient}.
The resolution of the final output is $0.5m$, resulting in $200\times200$ grid map.

\subsection{Training Settings}
All models are trained for 24 epochs with AdamW~\cite{loshchilov2018decoupled} optimizer in an end-to-end manner.
Image backbones are pre-trained on ImageNet~\cite{deng2009imagenet}.
In Table~\ref{table:appendix_settings}, we provide ResNet~\cite{he2016deep} and ConvNeXt~\cite{liu2022convnet} training settings used for our main results.

For image and radar data augmentation (in perspective view), we use resize, crop, and horizontal flipping augmentation following standard practices~\cite{huang2021bevdet, li2022bevdepth}.
We discard rotation augmentation since the rotation can have an adverse effect when collapsing the height dimension in radar-assisted view transformation (RVT).
Note that the same data augmentation is applied to the image and radar in the perspective view.

For BEV augmentation, we use random flipping along $X$ and $Y$ axis, global rotation between $[-\pi/8, \pi/8]$, and global scale between $[0.95, 1.05]$.
BEV data augmentation is applied to the BEV feature map and ground truth boxes correspondingly.
Note that ground-truth sampling augmentation (GT-AUG)~\cite{yan2018second} is not used in our experiments, and we leave GT-AUG for a multi-modal setting~\cite{chen2022autoalignv2, zhang2020exploring} as future work.

\setlength{\tabcolsep}{0.4em}
\begin{table}[!t]
\begin{center}
\resizebox{1.0\columnwidth}{!}{
\begin{tabular}{l||cc}
    \hline
    configs & ResNet-50/101 & ConvNeXt-B \\
    \hline
    optimizer       & AdamW & AdamW \\
    base learning rate         & 2e-4 & 1e-4 \\
    backbone learning rate     & 2e-4/1e-4 & 5e-5 \\
    weight decay    & 1e-4 & 1e-2 \\
    optimizer momentum & $\beta_1,\beta_2=0.9,0.999$ & $\beta_1,\beta_2=0.9,0.999$ \\
    batch size      & 64/32 & 16 \\
    training epochs & 24 & 24 \\
    lr schedule     & step decay & step decay \\
    gradient clip   & 5 & 5 \\
    stochastic depth \cite{huang2016deep} & None & 0.4 \\
    layer scale \cite{touvron2021going} & None & 1.0 \\
    \hline
\end{tabular}}
\end{center}
\vspace{-3pt}
\caption{
    Training settings for the main results.
}
\vspace{-8pt}
\label{table:appendix_settings}
\end{table}

\subsection{Baselines for Ablation Studies} \label{appendix:baselines}
We conduct three baselines BEVDepth~\cite{li2022bevdepth}, CenterPoint~\cite{yin2021center}, and BEVFusion~\cite{liu2022bevfusion} for camera-only, point-only, and camera-point fusion detectors.
Note that CenterPoint and BEVFusion originally take LiDAR points as input and we replace LiDAR points $(x, y, z, intensity)$ to radar points $(x, y, z, RCS, Doppler)$ without network modification. 

\renewcommand{\thefootnote}{1}
For BEVDepth, we use the official code\footnote{\url{https://github.com/Megvii-BaseDetection/BEVDepth}} without class-balanced grouping and sampling (CBGS)~\cite{zhu2019class} and exponential moving average (EMA).

\renewcommand{\thefootnote}{2}
For CenterPoint, we use MMDetection\footnote{\url{https://github.com/open-mmlab/mmdetection3d}} implementation using PointPillar~\cite{lang2019pointpillars} backbone with $(0.2, 0.2, 8) m$ pillar size.
Different from the official implementation, CBGS~\cite{zhu2019class} and GT-AUG~\cite{yan2018second} are discarded for fair comparisons.

For BEVFusion, we use BEVDepth for obtaining the camera BEV feature map and CenterPoint-Pillar for point BEV feature maps and fuse them by a single $3\times3$ convolution layer following official implementation.
Note that our BEVFusion may yield better performance since our implementation uses BEVDepth for the camera stream, while the original BEVFusion uses LSS~\cite{philion2020lift}.

\setlength{\tabcolsep}{0.3em}
\begin{table*}[!t]
\begin{center}
\resizebox{1.0\textwidth}{!}{
\begin{tabular}{l|c||cccccccccc|c}
    \hline
    Method & Input &Car&Truck&Bus&Trailer&C.V.&Ped.	&M.C.&Bicycle&T.C.&Barrier&mAP\\
    \hline
    CenterPoint-P \cite{yin2021center} & L & 83.9&49.5&61.9&34.1&12.3&76.9&44.1&18.0&54.0&59.1&49.4\\
    \hline
    CenterNet \cite{Zhou2019} & C & 48.4&23.1&34.0&13.1&3.5&37.7&24.9&23.4&55.0&45.6&30.6\\
    CenterFusion \cite{nabati2021centerfusion} & C+R & 52.4\scriptsize{(+4.0)}&26.5\scriptsize{(+3.4)}&36.2\scriptsize{(+2.2)}&15.4\scriptsize{(+2.3)}&5.5\scriptsize{(+2.0)}&
    38.9\scriptsize{(+1.2)}&30.5\scriptsize{(+5.6)}&22.9\scriptsize{(-0.5)}&56.3\scriptsize{(+1.3)}&47.0\scriptsize{(+1.4)}&33.2\scriptsize{(+2.6)}\\
    \hline
    CRAFT-I \cite{kim2022craft} & C & 52.4&25.7&30.0&15.8&5.4&39.3&28.6&29.8&57.5&47.8&33.2\\
    CRAFT \cite{kim2022craft} & C+R & 
    69.6\scriptsize{(\textbf{+17.2})}&37.6\scriptsize{(\textbf{+11.9})}&47.3\scriptsize{(\textbf{+17.3})}&20.1\scriptsize{(+4.3)}&10.7\scriptsize{(+5.3)}&
    46.2\scriptsize{(+6.9)}&39.5\scriptsize{(\textbf{+10.9})}&31.0\scriptsize{(+1.2)}&57.1\scriptsize{(-0.4)}&51.1\scriptsize{(+3.3)}&41.1\scriptsize{(+7.9)}\\
    \hline
    BEVDepth \cite{li2022bevdepth} & C & 55.3&25.2&37.8&16.3&7.6&36.1&31.9&28.6&53.6&55.9&34.8\\ 
    CRN & C+R & 
    73.6\scriptsize{(\textbf{+18.3})}&44.5\scriptsize{(\textbf{+19.3})}&55.6\scriptsize{(\textbf{+17.8})}&22.0\scriptsize{(+5.7)}&15.4\scriptsize{(+7.8)}&
    50.2\scriptsize{(\textbf{+14.1})}&54.7\scriptsize{(\textbf{+22.8})}&48.9\scriptsize{(\textbf{+20.3})}&61.4\scriptsize{(+7.8)}&63.8\scriptsize{(+7.9)}&49.0\scriptsize{(\textbf{+14.2})}\\
    \hline
\end{tabular}}
\end{center}
\vspace{-4pt}
\caption{
Per-class comparisons on nuScenes \texttt{val} set. 
`C.V.', `Ped.', `M.C.', and `T.C.' denote construction vehicle, pedestrian, motorcycle, and traffic cone, respectively.
CenterNet~\cite{Zhou2019}, CRAFT-I~\cite{kim2022craft}, and BEVDepth~\cite{li2022bevdepth} are camera baselines of CenterFusion~\cite{nabati2021centerfusion}, CRAFT~\cite{kim2022craft}, and CRN.
CenterPoint-P and BEVDepth results are from MMDetection3D and the official code.
}
\vspace{-6pt}
\label{table:appendix_perclass}
\end{table*}
\subsection{Details of Long-Range Model} \label{appendix:CRN_longrange}
To analyze the performance of CRN over long perception ranges, we increase the perception range of baselines to $[-102.4, 102.4] m$.
For camera streams, we increase the range of depth distribution from $[2.0, 58.0] m$ to $[2.0, 116.0] m$, and the number of depth bins is doubled to $D=224$.
For point streams, the range of point cloud and pillars are increased to correspond to the perception range.
Note that we use the same pillar size $(0.2, 0.2) m$ and downsampling factor of 4, resulting in a $256\times 256$ BEV feature map for all baselines.

For training and evaluating long-range models, we increase the `class range' in nuScenes~\cite{Caesar2020} twice to filter the ground truth and predictions.
Particularly, the class range of car, truck, bus, trailer, and construction vehicle are 100$m$, pedestrian, motorcycle, and bicycle are 80$m$, traffic cone and barrier are 60$m$.
Moreover, nuScenes filters annotation that does not contain at least single LiDAR or radar point inside the 3D bounding box (denote as `points in box filtering') for training and evaluation, but we disable this filtering for thorough analysis.
Thus, some moving objects are visible on the image but do not have annotations (due to not enough points to label), and some static objects can have annotations but are not visible on the image (labeled on the previous timestamp but occluded on the current timestamp) in our setting.
Although disabling point filtering may cause inconsistency between input data and annotation and harms performance during training, all methods are trained and evaluated using the same setting for a fair comparison.
We find that the inference speed of CenterPoint~\cite{yin2021center} with radar input is much faster than LiDAR input, assuming that the sparsity of radar points can highly benefit from voxelization and sparse convolution~\cite{yan2018second}.

\section{Additional Experimental Results} \label{appendix:experiment}
\subsection{Per-Class Analysis}
In Table~\ref{table:appendix_perclass}, we compare the performance improvement of camera-radar methods over camera-only baselines.
For fair comparisons, we report $256\times 704$ and R50 models for BEVDepth and CRN.
Corresponds to results on CRAFT~\cite{kim2022craft}, metallic and frequently appeared on road classes (car, truck, bus, and motorcycle) gain significant improvements.
Different from CRAFT, ours also shows a huge improvement in non-metallic classes (pedestrian, bicycle, traffic cone, and barrier).
Moreover, we find that the performance gain of using radar is much more significant on ours than other fusion methods.
Considering the performance difference of camera baselines are not significant, results in Table~\ref{table:appendix_perclass} demonstrate that the design of fusion methods greatly affects the performance.

\setlength{\tabcolsep}{0.5em}
\begin{table}[!t]
\begin{center}
\resizebox{0.8\columnwidth}{!}{
\begin{tabular}{c||cc|ccc}
    \hline
    \# Frames & NDS & mAP & mATE & mAOE & mAVE \\
    \hline
    1 & 50.3 & 42.9 & 0.519 & 0.577 & 0.520\\
    2 & 54.5 & 46.0 & 0.495 & 0.538 & 0.350 \\
    3 & 55.7 & 47.3 & 0.480 & 0.507 & 0.342 \\
    \gr 4 & 56.0 & 48.1 & 0.474 & 0.541 & 0.328 \\
    5 & 56.4 & 48.4 & 0.469 & 0.515 & 0.345 \\
    \hline
\end{tabular}}
\end{center}
\vspace{-3pt}
\caption{
    Ablation of temporal frames.
}
\vspace{-4pt}
\label{table:appendix_frames}
\end{table}

\subsection{Design Decisions} \label{appendix:design}
We study architecture designs that affect the performance of CRN to provide insights on the proposed sensor fusion framework.

\vspace{3pt}
\noindent\textbf{Temporal Frames.\hspace{0.2cm}}
We accumulate multiple BEV feature maps on channel dimension by concatenation and aggregate them by a few convolutional layers before feeding them to the BEV backbone.
We find that the time interval of 1 second yields a better performance than 0.5 second proposed in previous approaches~\cite{li2022bevformer, li2022bevdepth}.
Compared to temporal stereo methods~\cite{li2022bevstereo, park2022time}, ours does not require sequential data input for obtaining the BEV feature map; thus, using an arbitrary number of BEV feature maps does not increase latency.
We note that BEV feature maps on previous timestamps are obtained without gradients during training following standard practices.

As shown in Table~\ref{table:appendix_frames}, using multiple temporal frames significantly improves mAP, mATE, and mAVE.
Corresponding to results on recent approaches using temporal BEV feature maps~\cite{park2022time}, a larger number of frames consistently yields better performance.
However, we observe the unstable orientation error (mAOE), suggesting room for improvement in utilizing BEV feature maps, and we leave this as future work.
As the performance gain is saturated on four frames, we decide to use four frames considering computation time and memory during training.

\vspace{3pt}
\noindent\textbf{Sparse Aggregation.\hspace{0.2cm}}
In Table~\ref{table:appendix_sparse}, we ablate the number of $N_k$ feature grids on sparse aggregation settings.
Note that the total number of BEV feature grids is $N = 256\times 256 = 65536$ in our long-range setting and we report the performance on \textit{Car} class at 100$m$ perception range.
Since the computational complexity of sparse aggregation $\mathcal{O}(2N_k+N_kK)$ is linear to sparse input queries $N_k$, using a small set of features for MFA significantly reduces the computation of Multi-modal Deformable Cross Attention (MDCA).
More specifically, using 4096 size queries reduce the latency of MFA by 76.4\% (21.01$ms$ to 4.96$ms$) on $256\times 256$ size BEV grid.
However, as the BEV feature map becomes sparse and discretized after top-k sampling, the performance is degraded.
We find that the performance drops on True Positive metrics (\textit{e.g.}, ATE, AOE, AVE) are more significant than AP, assuming that the classification network can maintain its performance, but the regression network suffers from sparsely spread BEV features to regress objects' attributes.

\setlength{\tabcolsep}{0.75em}
\begin{table}[!t]
\begin{center}
\resizebox{0.82\columnwidth}{!}{
\begin{tabular}{c||c|ccc|c}
    \hline
    \# Top-K & AP & ATE & AOE & AVE & FPS \\
    \hline
    1024  & 49.8 & 0.399 & 0.216 & 0.371 & 14.1 \\
    2048  & 52.4 & 0.382 & 0.202 & 0.352 & 14.0 \\
    \gr 4096  & 54.0 & 0.367 & 0.194 & 0.340 & 14.0 \\
    8192  & 54.6 & 0.362 & 0.191 & 0.352 & 13.8 \\
    All & 56.9 & 0.325 & 0.158 & 0.298 & 11.5 \\
    \hline
\end{tabular}}
\end{center}
\vspace{-3pt}
\caption{
    Ablation of sparse aggregation.
}
\vspace{-8pt}
\label{table:appendix_sparse}
\end{table}

\section{Additional Qualitative Results} \label{appendix:qualitative}
We show additional qualitative results of 3D object detection (long-range $256\times 704$ and R50 model) and BEV segmentation ($256\times 704$ and R50 model).

\begin{figure*}[t]
\begin{center}
\includegraphics[width=0.9\textwidth]{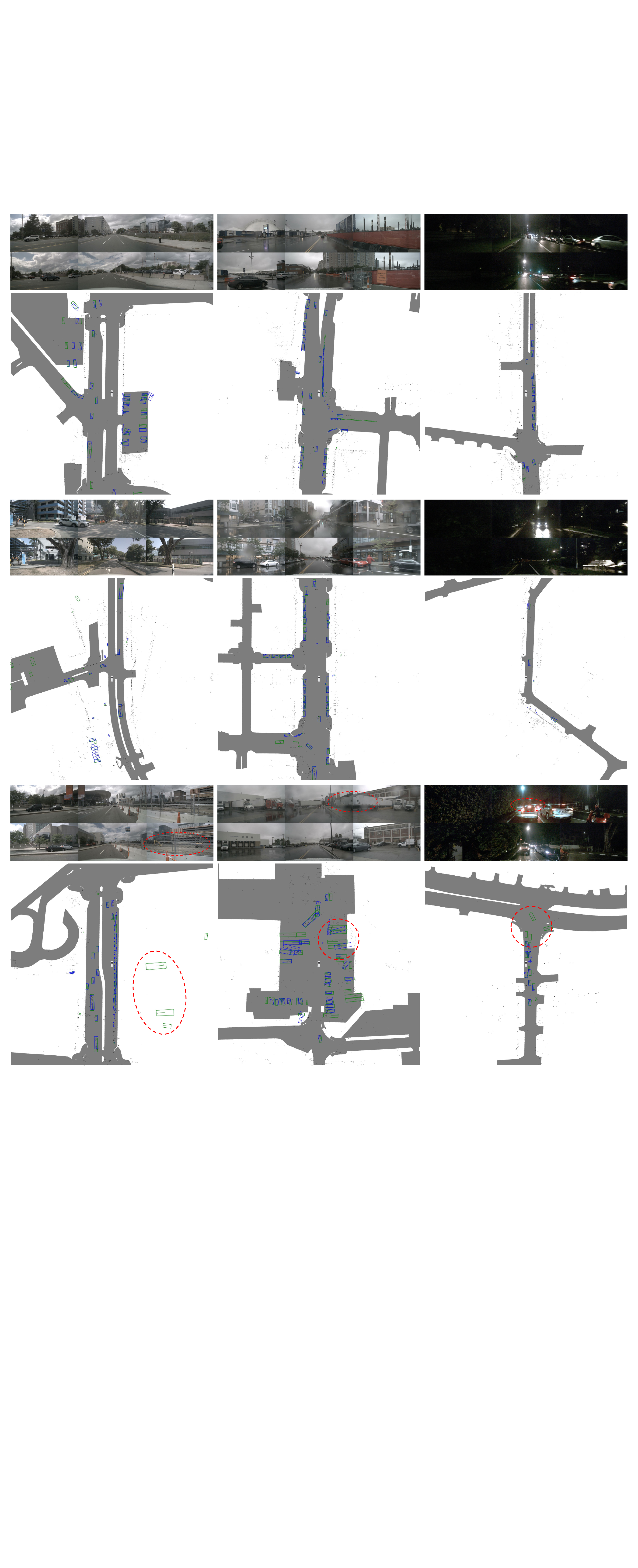}
\end{center}
\vspace{-10pt}
\caption{\small
Additional qualitative results of 3D object detection on nuScenes \texttt{val} set: from left to right, day, rainy, and night scenarios.
Green boxes are ground truths, blue boxes are our prediction results, and black dots are radar points.
We also show the failure cases and highlight them with red circles on the bottom row.
Ground truth maps on the background are used for visualization.
Best viewed in color with zoom in.
}
\label{fig:qualitative_3ddet}
\end{figure*}

To visualize 3D detection results for the range of $200m\times200m$, we disable points in box filtering as described in Appendix~\ref{appendix:CRN_longrange}.
As can be seen in Fig.~\ref{fig:qualitative_3ddet}, CRN is capable of detecting objects even at a very far distance under various and complex driving scenarios.
Thanks to radar fusion, objects strongly occluded by other objects or hardly visible by low lighting are successfully detected by ours.
Moreover, even if some objects do not have radar point returns, CRN can still detect them by image only.
Failure cases of CRN are likely caused when objects are rare classes and do not without radar points (\textit{e.g.}, construction vehicles behind wire mesh or trailers heavily occluded).

\begin{figure*}[t]
\begin{center}
\includegraphics[width=1.0\textwidth]{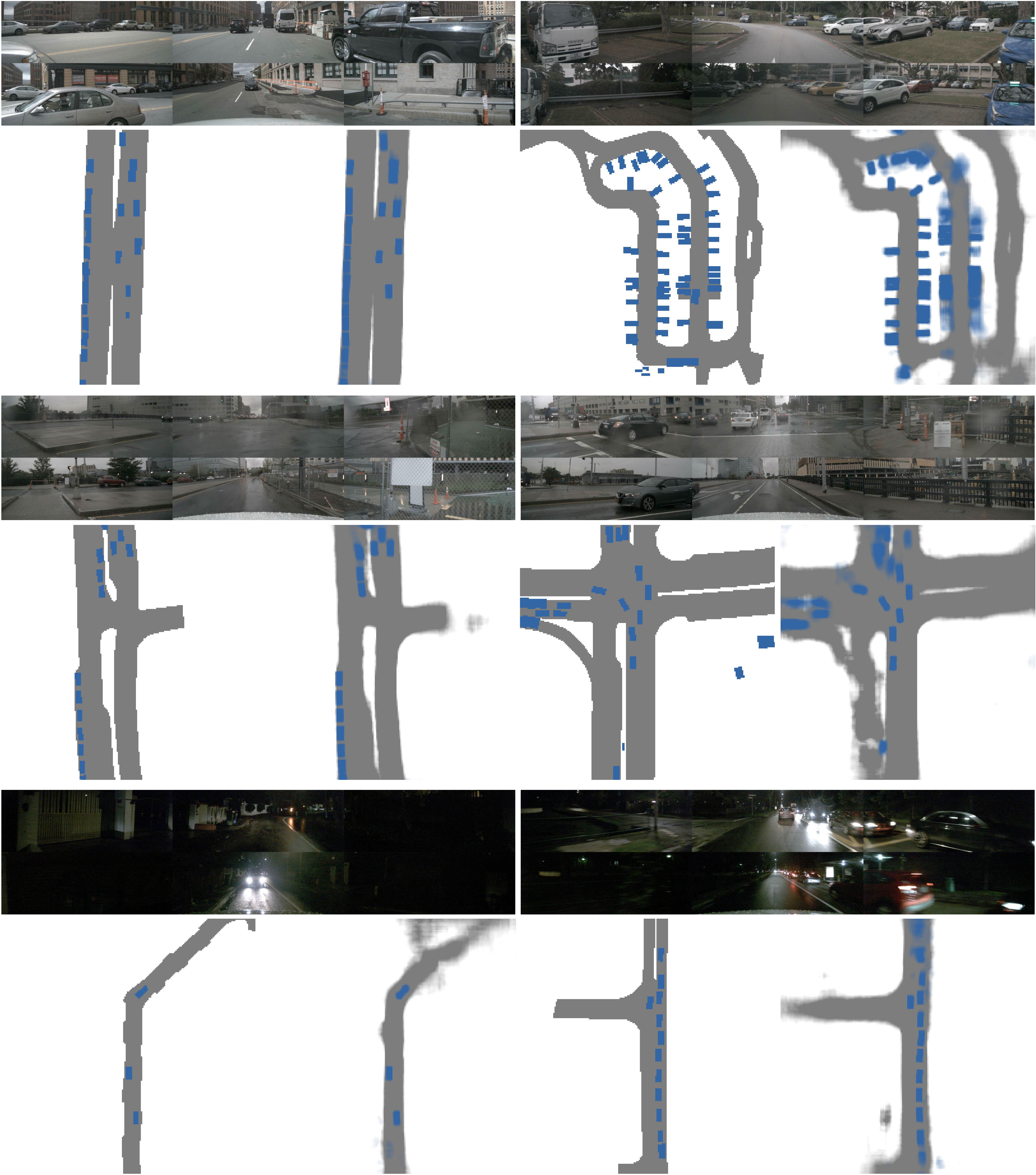}
\end{center}
\caption{\small
Qualitative results of BEV segmentation on nuScenes \texttt{val} set on various road shapes and weather conditions: from top to bottom, day, rainy, and night scenarios.
Images on the top are the six camera views surrounding the vehicle, the bottom left is ground truth, and the bottom right is our prediction results.
Best viewed in color with zoom in.
}
\label{fig:qualitative_bevseg}
\end{figure*}

We further visualize BEV segmentation results in the range of $100m\times100m$, following the same setting of previous works.
As shown in Fig.~\ref{fig:qualitative_bevseg}, CRN is also capable of accurately predicting segmentation occupancy of drivable region and vehicle.
Thanks to our camera-radar fusion framework to generate a semantically rich and spatially accurate feature map, our results show stable performance under various lighting and weather conditions.
CRN can further predict occupancy of the vehicles with a complete shape both at nearby and faraway distances, even when the vehicles are partially visible.
In terms of the drivable region, CRN can successfully predict the complex shapes of the road even under occlusions.

\end{document}